\documentclass[11pt]{article}

\usepackage[margin=1in]{geometry}
\usepackage{times}
\usepackage{latexsym}
\usepackage{microtype}
\usepackage{amsmath,amssymb}
\usepackage{pifont}
\newcommand{\cmark}{\ding{51}}
\newcommand{\xmark}{\ding{55}}
\usepackage{colortbl}
\usepackage{booktabs}
\usepackage{graphicx}
\usepackage{pgfplots}
\pgfplotsset{compat=1.18}
\usepgfplotslibrary{dateplot}
\usepackage{caption}
\usepackage{enumitem}
\usepackage{placeins}
\usepackage{float}
\usepackage{url}
\usepackage{kotex}
\usepackage{fvextra}
\usepackage[many,listings]{tcolorbox}
\usepackage{subcaption}
\usepackage[numbers]{natbib}
\usepackage[colorlinks=true,allcolors=blue]{hyperref}

\definecolor{main}{HTML}{5989cf}
\definecolor{sub}{HTML}{cde4ff}
\definecolor{ssgreen}{HTML}{228B22}

\tcbset{
    sharp corners,
    colback = white,
    before skip = 0.2cm,
    after skip = 0.5cm
}

\newtcolorbox{promptbox}{
    colback = sub,
    colframe = main,
    boxrule = 0pt,
    leftrule = 6pt,
    fontupper = \ttfamily\small
}

\newtcolorbox{codebox}{
    colback = sub,
    colframe = main,
    boxrule = 0pt,
    toprule = 3pt,
    bottomrule = 3pt,
    fontupper = \ttfamily\small
}

\newcommand{\micon}[1]{\raisebox{-0.15em}{\includegraphics[height=0.8em]{assets/icons/#1}}}
\newcommand{\mname}[1]{#1}

\title{KorMedMCQA-V: A Multimodal Benchmark for Evaluating Vision-Language Models on the Korean Medical Licensing Examination}

\author{Byungjin Choi\textsuperscript{1,*} \quad
Seongsu Bae\textsuperscript{2,*} \quad
Sunjun Kweon\textsuperscript{2} \quad
Edward Choi\textsuperscript{2,\textdagger}\\
\\
\textsuperscript{1}Ajou University School of Medicine\\
\textsuperscript{2}KAIST
}

\date{}

\begin{document}
\maketitle
\thispagestyle{empty}
{\let\thefootnote\relax\footnotetext{\textsuperscript{*}Co-first authors. \quad \textsuperscript{\textdagger}Corresponding author.}}

\begin{abstract}
	We introduce \textbf{KorMedMCQA-V}, a Korean medical licensing-exam-style multimodal multiple-choice question answering benchmark for evaluating vision-language models (VLMs).
	The dataset consists of 1,534 questions with 2,043 associated images from Korean Medical Licensing Examinations (2012--2023), with about 30\% containing multiple images requiring cross-image evidence integration.
	Images cover clinical modalities including X-ray, computed tomography (CT), electrocardiography (ECG), ultrasound, endoscopy, and other medical visuals.
	We benchmark over 50 VLMs across proprietary and open-source categories---spanning general-purpose, medical-specialized, and Korean-specialized families---under a unified zero-shot evaluation protocol.
	The best proprietary model (Gemini-3.0-Pro) achieves 96.9\% accuracy, the best open-source model (Qwen3-VL-32B-Thinking) 83.7\%, and the best Korean-specialized model (VARCO-VISION-2.0-14B) only 43.2\%.
	We further find that reasoning-oriented model variants gain up to +20 percentage points over instruction-tuned counterparts, medical domain specialization yields inconsistent gains over strong general-purpose baselines, all models degrade on multi-image questions, and performance varies notably across imaging modalities.
	By complementing the text-only KorMedMCQA benchmark, KorMedMCQA-V forms a unified evaluation suite for Korean medical reasoning across text-only and multimodal conditions.
	The dataset is available via Hugging Face Datasets: \href{https://huggingface.co/datasets/seongsubae/KorMedMCQA-V}{https://huggingface.co/datasets/seongsubae/KorMedMCQA-V}.
\end{abstract}


\section{Introduction}

Large language models (LLMs) and vision-language models (VLMs) have advanced medical question answering and image understanding.
National medical licensing examinations—standardized assessments that every practicing physician must pass—offer a particularly reliable source for benchmarking clinical competency, and text-based benchmarks derived from such exams now span multiple countries and languages (\textit{e.g.}, MedQA~\cite{jin2021disease}, MedMCQA~\cite{pal2022medmcqa}, CMExam~\cite{liu2023benchmarking}).

However, real licensing examinations routinely require interpreting visual evidence—radiographs, pathology slides, ECGs—making multimodal reasoning integral to clinical assessment.
While multimodal exam-style benchmarks have recently emerged for several countries (\textit{e.g.}, WorldMedQA-V~\cite{matos2025worldmedqa}, KokushiMD-10~\cite{liu2025kokushimd}, PerMed-MM~\cite{khoramfar2025permed}, MMMED~\cite{riccio2025multilingual}), no such resource exists for the Korean Medical Licensing Examination.
KorMedMCQA~\cite{kweon2024kormedmcqa} introduced Korean medical multiple-choice question answering (MCQA) but covers only text-only questions, leaving the image-based portion of the exam unaddressed.

\begin{figure}[t]
	\centering
	\footnotesize
	\setlength{\fboxsep}{10pt}
	\fbox{\begin{minipage}{0.96\linewidth}
		\begin{minipage}{0.30\linewidth}
			\centering
			\includegraphics[width=\linewidth]{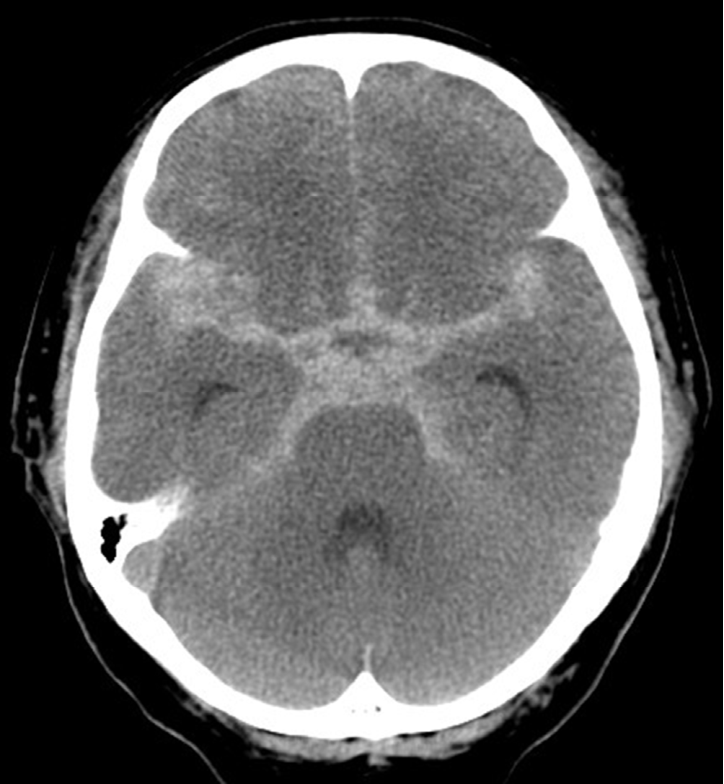}
		\end{minipage}%
		\hfill
		\begin{minipage}{0.65\linewidth}
			{\footnotesize \textbf{Question:}} 67세 남자가 6시간 전부터 갑자기 머리가 깨질 듯이 아파서 응급실에 왔다. 뒷목이 당기고, 멍한 느낌이 들었고, 어지럽고 메스꺼워서 토했다. 혈압강하제를 복용 중이다. 혈압 183/84 mmHg, 맥박 51회/분, 호흡 18회/분, 체온 36.5$^\circ$C이다. 의식은 명료하다. 뇌 컴퓨터단층촬영 사진이다. 진단은? \textit{(A 67-year-old man presents to the emergency room with a 6-hour history of sudden-onset severe headache. He reports neck stiffness, mental fogginess, dizziness, and vomiting. He is on antihypertensive medication. Vital signs: BP 183/84 mmHg, pulse 51/min, RR 18/min, temperature 36.5$^\circ$C. He is alert. Brain CT is shown. What is the diagnosis?)}
			\begin{enumerate}[label=\textbf{\Alph*.}, leftmargin=*, itemsep=0pt, topsep=1pt]
				\item 뇌내혈종 \textit{(Intracerebral hematoma)}
				\item 경막외혈종 \textit{(Epidural hematoma)}
				\item 경막밑혈종 \textit{(Subdural hematoma)}
				\item 뇌실내출혈 \textit{(Intraventricular hemorrhage)}
				\item 거미막밑출혈 \textit{(Subarachnoid hemorrhage)}
			\end{enumerate}
		\end{minipage}
	\end{minipage}}
	\caption{Representative KorMedMCQA-V multiple-choice item with an associated medical image (imaging modality: CT). English translations are provided in parenthesized italics for readability. Models are given the full question stem, all answer choices, and the image(s), and must output a single option label (A--E).}
	\label{fig:example_mcqa}
\end{figure}

To address these gaps, we introduce KorMedMCQA-V, a Korean medical licensing examination-style multimodal multiple-choice question answering benchmark.
KorMedMCQA-V contains 1,534 questions with images from Korean licensing exams, with roughly 70\% having a single image and 30\% having multiple images, reflecting common multi-panel exam formats.
For example, Figure~\ref{fig:example_mcqa} shows a representative item requiring diagnosis from a brain CT.
We evaluate VLMs spanning general-purpose, medical-specialized, and Korean-specialized families under a unified zero-shot protocol, analyzing performance by image modality, model type, and single- vs.~multi-image settings.

\FloatBarrier

Our contributions are:
(1) We introduce KorMedMCQA-V, a Korean licensing-exam-style multimodal MCQA benchmark with image-based questions.
(2) We benchmark diverse VLM families (general-purpose, medical-specialized, and Korean-specialized) under a unified protocol.
(3) We provide factorized analyses by modality, model type, and single- vs.~multi-image settings to identify bottlenecks in Korean multimodal medical reasoning.

\section{Related Work}

\subsection{Medical Licensing Exam Benchmarks}
Text-only medical examination benchmarks are widely used to evaluate the medical knowledge and reasoning capabilities of LLMs.
MedQA~\cite{jin2021disease} is the most commonly used USMLE-style benchmark, and CMExam~\cite{liu2023benchmarking} similarly targets the Chinese National Medical Licensing Examination.
Multimodal exam-style benchmarks such as WorldMedQA-V~\cite{matos2025worldmedqa}, KokushiMD-10~\cite{liu2025kokushimd}, PerMed-MM~\cite{khoramfar2025permed}, and MMMED~\cite{riccio2025multilingual} have further extended evaluation to image-based medical questions.
KorMedMCQA~\cite{kweon2024kormedmcqa} established a text-only benchmark for Korean Medical Licensing Examinations, and our KorMedMCQA-V extends coverage to image-based questions from the same examinations.

\subsection{Korean Vision-Language Benchmarks}
The development of Korean vision-language models~\cite{cha2025varco,yoo2024hyperclova,skt2025axvl,bak2025kanana} has been accompanied by benchmarks that evaluate diverse VLM capabilities.
Existing resources cover general visual question answering~\cite{kim2025koffvqa}, text-rich understanding and OCR~\cite{hwang2025kreta,lee2025kocrbench}, visual document retrieval~\cite{lee2025sds}, cultural-contextual interpretation~\cite{park2025evaluating}, and model robustness under ambiguous queries~\cite{choi2026users}.
However, these benchmarks are concentrated in the general domain; domain-specific evaluation, particularly in medicine where visual evidence such as radiographs, pathology slides, and ECGs is integral to clinical reasoning, has received little attention for Korean.
KorMedMCQA-V addresses this gap as the first Korean exam-style multimodal benchmark for medical licensing examination questions.

\section{Benchmark Construction}

\subsection{Data Acquisition and Selection}
KorMedMCQA-V is sourced from the official Korean Medical Licensing Examination (KMLE) administered between 2012 and 2023.
We retain only items that contain one or more images, focusing on the doctor licensing examinations.
A preliminary analysis showed that other professional tracks (\textit{i.e.}, nurse, pharmacist, dentist) contain fewer than 5\% image-based questions, making them unsuitable for a robust multimodal benchmark.
We further exclude R-type questions (\textit{i.e.}, extended matching items with 8--10 options where examinees must select the specified number of correct answers) and unreleased items to maintain evaluation consistency.

\subsection{Data Extraction}
We develop a systematic pipeline to extract images and text directly from the official PDF sources.
We use PyMuPDF\footnote{\url{https://github.com/pymupdf/PyMuPDF}} to extract embedded images from each page.
Images and their corresponding questions are linked by locating picture number labels (\textit{i.e.}, figure/picture identifiers) near the image and matching them to references in the question stem via regular expressions.
All automatic extractions are manually verified and corrected where necessary.

\subsection{Imaging Modality Annotation}
\label{sec:modality_annotation}
To annotate image modality, we use a consensus-based pipeline with four recent VLMs: Gemini-3.0-Flash~\cite{google2025gemini3}, GLM-4.6V-Flash and GLM-4.6V~\cite{zeng2025glm}, and Qwen3-VL-30B-A3B-Thinking~\cite{yang2025qwen3}.
Each model assigns one label from a fixed set of nine categories (XRAY, CT, MRI, US, ENDOSCOPY, ECG, NST, PBS, OTHER\footnote{The OTHER category covers clinical photographs, diagrams, and charts.}).
The model consensus serves as an initial annotation to streamline manual verification; all labels are then reviewed by a clinician, with unanimous cases (89.2\% of 2,043 image instances) verified efficiently and the remaining 220 disagreement cases (10.8\%) adjudicated through more detailed examination.
The exact prompt template used for model-based modality annotation is provided in Appendix~\ref{app:modality_prompt}.

\subsection{Dataset Overview}
\label{sec:dataset_overview}
KorMedMCQA-V comprises 1,534 multimodal questions from the 2012--2023 doctor licensing examinations, each requiring joint reasoning over a clinical scenario and one or more medical images.
Question stems average 50 words (median: 48), reflecting the detailed patient scenarios typical of licensing exams.
Most questions contain a single image (69.7\%), while a notable portion include multiple images (30.3\%), most of which contain exactly two (27.8\% of all questions), mirroring clinical cases that require integrating multiple views or modalities.
The images span nine medical imaging modalities: X-ray (28.7\%), clinical photographs and diagrams (27.1\%), CT (16.4\%), ECG (8.0\%), US (6.8\%), Endoscopy (6.0\%), NST (2.6\%), PBS (2.4\%), and MRI (2.0\%).
Image resolutions vary widely, ranging from $213\times129$ to $4252\times4795$ pixels (median: $1098\times890$).

\section{Experiments}
\subsection{Experimental Setup }
We formulate KorMedMCQA-V as a 5-way multiple-choice question answering task: each example consists of a question stem, five answer options, and one or more associated images; models must select the correct option label (A--E) using both textual and visual evidence.
We evaluate all models in a zero-shot, closed-book setting (\textit{i.e.}, no external tools) using a single prompt template that instructs the model to output the selected label in JSON format; we robustly parse outputs to allow minor formatting variations.

For image preprocessing, we use each model's default processor; for multi-image examples, all images are provided in their original exam order.
For generation hyperparameters (\textit{e.g.}, temperature, top\_p), we use each model's default or recommended settings unless otherwise specified.
For open-source models, we run inference using HF Transformers~\cite{wolf2019huggingface} or vLLM~\cite{kwon2023efficient} with each model's official chat template and processor.

We score predictions by exact match between the predicted and gold option labels and report accuracy.
For open-source models, we run three random seeds and report the average.
Full reproducibility details (model versions, prompts, and hyperparameters) are provided in the Reproducibility section.

\subsection{Models }
To disentangle the effects of scale, reasoning-oriented training, domain adaptation, and Korean language coverage, we evaluate a broad set of vision-language models (VLMs) spanning diverse parameter scales and training paradigms.
We organize them into proprietary and open-source categories; open-source models are further divided into general-purpose, medical-specialized, and Korean-specialized groups.
The complete model list with exact identifiers and configurations is provided in Appendix~\ref{app:models}.

We establish two reference points to contextualize model performance.
First, we include a majority-label baseline that always predicts the most frequent ground-truth option label (A--E) to quantify option-label frequency bias (\textit{i.e.}, answer-position bias) and provide a minimal lower bound.
Second, we report results from proprietary models as high-capacity reference points, including OpenAI models (GPT-5 and GPT-5-mini~\cite{openai2026gpt5}) and Google Gemini (Gemini-3.0-Pro and Gemini-3.0-Flash~\cite{google2025gemini3}).

General-purpose VLMs include the InternVL 3.5 series~\cite{wang2025internvl3}, Qwen2.5-VL~\cite{bai2025qwen25vl} and Qwen3-VL series~\cite{yang2025qwen3}, Gemma 3~\cite{team2025gemma}, Kimi-VL~\cite{kimiteam2025kimivl}, GLM-4.6V~\cite{zeng2025glm}, and Ministral-3~\cite{liu2026ministral3}.
To probe the impact of explicit reasoning, we additionally evaluate both instruction-tuned and reasoning-oriented variants for families offering paired checkpoints (\textit{e.g.}, Qwen3-VL Instruct/Thinking, Ministral-3 Instruct/Reasoning).

To test whether domain adaptation helps on exam-style multimodal medical reasoning, we evaluate medical-specialized VLMs trained on biomedical corpora or medical VQA data: MedGemma~\cite{sellergren2025medgemma} (built on Gemma 3), Lingshu~\cite{xu2025lingshu} (built on Qwen2.5-VL), and Hulu-Med~\cite{jiang2025hulu} (built on Qwen2.5 and Qwen3).
To assess the impact of Korean language coverage, we include Korean-specialized VLMs pre-trained or fine-tuned on Korean corpora: VARCO-VISION-2.0~\cite{cha2025varco}, A.X-4.0-VL-Light~\cite{skt2025axvl}, HyperCLOVAX-SEED-Vision~\cite{yoo2024hyperclova}, and Kanana-1.5-V~\cite{bak2025kanana}.

\clearpage

\subsection{KorMedMCQA-V Results}
\begin{table}[t!]
	\centering
	\scriptsize
	\setlength{\tabcolsep}{3pt}
	\renewcommand{\arraystretch}{0.92}
	\caption{Main experimental results on KorMedMCQA-V (selected models). We report overall accuracy, per-modality accuracy, and accuracy by image count (1/2/3+ images). \textbf{Bold} indicates the best score within each column for each model category. The average row is computed over all 51 evaluated models. Parenthesized numbers below column headers indicate the number of image instances for modality columns and the number of questions for image count columns. Full results are available in Appendix~\ref{app:full_results}.}
	\label{tab:results_main_summary}
	\resizebox{\linewidth}{!}{%
		\begin{tabular}{lccccccccccccc}
			\toprule
			\textbf{Model}                                                     & \textbf{Overall} & \multicolumn{9}{c}{\textbf{Modality}} & \multicolumn{3}{c}{\textbf{\# Images}}                                                                                                                                                     \\
			\cmidrule(lr){3-11} \cmidrule(lr){12-14}
			                                                                   &                  & \textbf{XRAY}                         & \textbf{CT}                            & \textbf{ECG} & \textbf{US} & \textbf{Endo.} & \textbf{NST} & \textbf{PBS} & \textbf{MRI} & \textbf{Other} & \textbf{1} & \textbf{2} & \textbf{3+} \\
			                                                                   &                  & \scriptsize{(586)} & \scriptsize{(336)} & \scriptsize{(164)} & \scriptsize{(138)} & \scriptsize{(122)} & \scriptsize{(54)} & \scriptsize{(49)} & \scriptsize{(40)} & \scriptsize{(554)} & \scriptsize{(1,069)} & \scriptsize{(426)} & \scriptsize{(39)} \\
			\midrule
			\textit{Average (n=51)} & 55.9 & 55.2 & 51.5 & 58.5 & 54.5 & 52.9 & 50.5 & 57.7 & 59.0 & 59.2 & 57.0 & 53.8 & 50.3 \\
			\mname{Always choose majority label (E)} & 22.4 & 23.5 & 21.3 & 12.2 & 25.5 & 19.6 & 30.4 & 23.3 & 25.9 & 22.7 & 23.2 & 20.4 & 23.1 \\
			\midrule
			\multicolumn{14}{l}{\emph{Proprietary models}} \\
			\midrule
			\micon{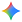}~\mname{Gemini-3.0-Pro} & \textbf{96.9} & \textbf{97.0} & \textbf{97.9} & \textbf{95.9} & \textbf{97.2} & 93.5 & 91.3 & \textbf{100.0} & \textbf{100.0} & 97.4 & \textbf{97.7} & \textbf{96.0} & 87.2 \\
			\micon{gemini.png}~\mname{Gemini-3.0-Flash} & \textbf{96.9} & 96.2 & \textbf{97.9} & 93.9 & 96.2 & \textbf{94.6} & \textbf{93.5} & \textbf{100.0} & \textbf{100.0} & \textbf{98.3} & \textbf{97.6} & 95.5 & \textbf{92.3} \\
			\micon{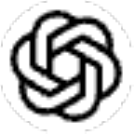}~\mname{GPT-5-2025-08-07} & 93.9 & 93.6 & 89.1 & 89.8 & \textbf{97.2} & 92.4 & 87.0 & \textbf{100.0} & \textbf{100.0} & 96.3 & 94.7 & 92.0 & \textbf{92.3} \\
			\micon{openai.pdf}~\mname{GPT-5-mini-2025-08-07} & 90.1 & 89.1 & 85.4 & 85.7 & 90.6 & 88.0 & 89.1 & 97.7 & 96.3 & 93.3 & 91.3 & 87.3 & 87.2 \\
			\midrule
			\multicolumn{14}{l}{\emph{General-purpose VLMs}} \\
			\midrule
			\micon{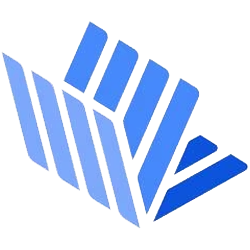}~\mname{InternVL3.5-4B} & 43.3 & 43.9 & 41.0 & 44.9 & 43.7 & 40.6 & 38.4 & 47.3 & 42.0 & 43.9 & 44.2 & 41.6 & 35.9 \\
			\micon{intervl.png}~\mname{InternVL3.5-8B} & 47.4 & 44.8 & 44.3 & 53.6 & 49.1 & 41.8 & 42.4 & 45.3 & 61.1 & 50.6 & 49.0 & 44.7 & 32.1 \\
			\micon{intervl.png}~\mname{InternVL3.5-38B} & 67.2 & 66.7 & 62.2 & 72.8 & 61.9 & 64.1 & 50.7 & 76.0 & 79.0 & 70.6 & 68.6 & 64.2 & 61.5 \\
			\micon{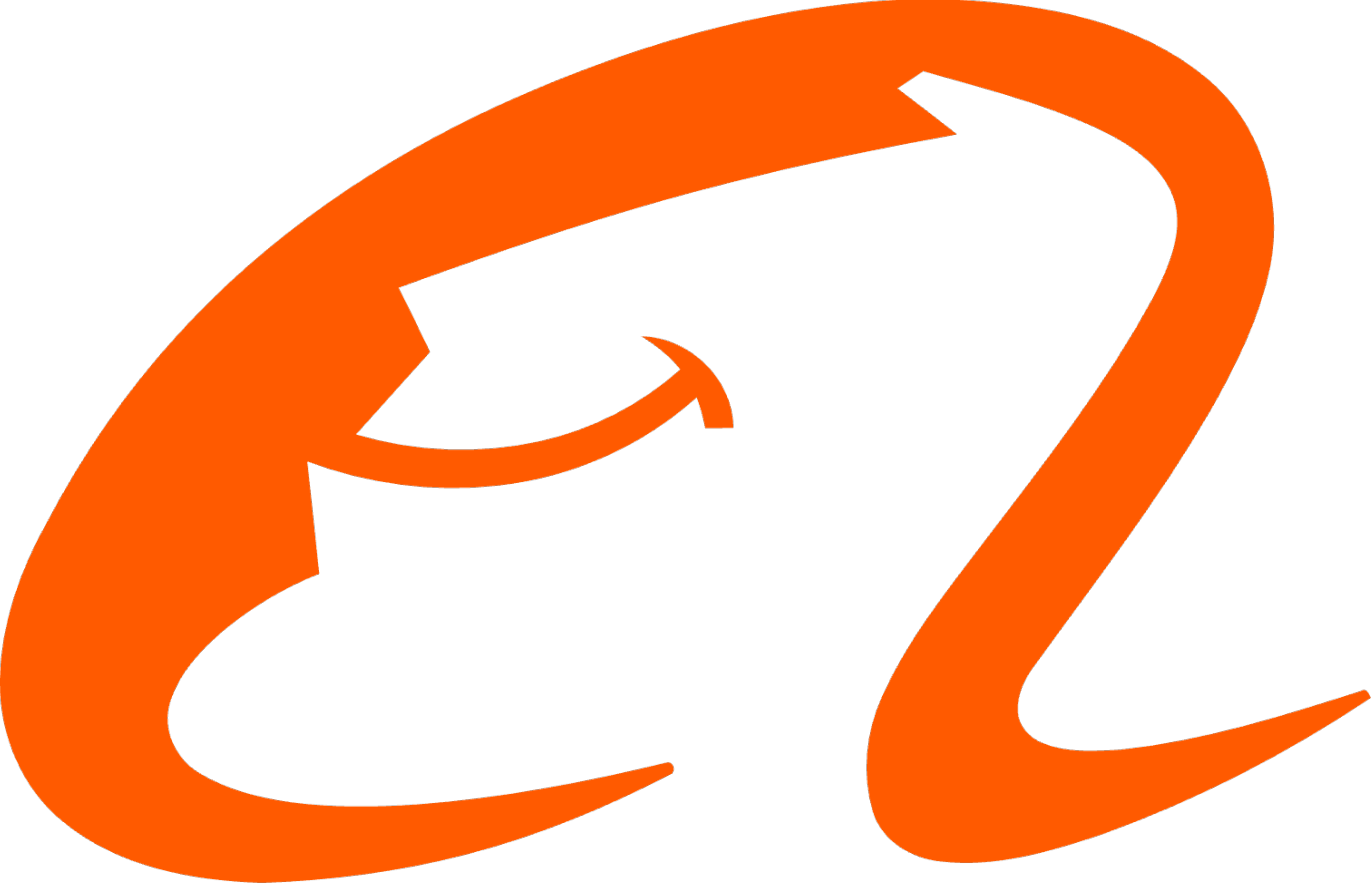}~\mname{Qwen2.5-VL-7B-Instruct} & 46.6 & 46.9 & 42.7 & 50.0 & 41.5 & 42.8 & 43.5 & 39.5 & 44.4 & 50.1 & 46.8 & 45.4 & 52.1 \\
			\micon{qwen.pdf}~\mname{Qwen2.5-VL-32B-Instruct} & 63.0 & 62.3 & 56.9 & 71.8 & 64.5 & 59.1 & 52.2 & 67.4 & 70.4 & 65.2 & 64.0 & 61.4 & 54.7 \\
			\micon{qwen.pdf}~\mname{Qwen3-VL-4B-Instruct} & 45.7 & 41.7 & 42.0 & 53.1 & 41.2 & 47.5 & 46.4 & 46.5 & 35.8 & 50.9 & 47.1 & 43.3 & 32.5 \\
			\micon{qwen.pdf}~\mname{Qwen3-VL-4B-Thinking} & 65.9 & 63.1 & 64.1 & 70.4 & 68.2 & 66.3 & 46.4 & 82.2 & 60.5 & 68.7 & 68.1 & 61.5 & 54.7 \\
			\micon{qwen.pdf}~\mname{Qwen3-VL-8B-Instruct} & 53.8 & 50.4 & 46.5 & 56.8 & 55.3 & 51.1 & 50.7 & 46.5 & 55.6 & 60.8 & 55.9 & 49.1 & 47.9 \\
			\micon{qwen.pdf}~\mname{Qwen3-VL-8B-Thinking} & 74.2 & 75.1 & 67.9 & 74.8 & 76.7 & 69.6 & 68.8 & 84.5 & 72.8 & 75.8 & 75.7 & 71.4 & 65.8 \\
			\micon{qwen.pdf}~\mname{Qwen3-VL-32B-Instruct} & 76.5 & 75.9 & 71.9 & 83.3 & 77.0 & 67.8 & 64.5 & 86.0 & 85.2 & 79.1 & 77.9 & 74.5 & 61.5 \\
			\micon{qwen.pdf}~\mname{Qwen3-VL-32B-Thinking} & \textbf{83.7} & \textbf{83.9} & \textbf{78.0} & \textbf{86.4} & \textbf{81.1} & \textbf{76.4} & \textbf{71.7} & \textbf{95.3} & \textbf{88.9} & \textbf{87.0} & \textbf{85.2} & \textbf{80.5} & 76.9 \\
			\micon{gemini.png}~\mname{Gemma-3-4B-IT} & 34.3 & 32.7 & 27.6 & 36.7 & 39.6 & 26.1 & 30.4 & 27.9 & 48.1 & 38.7 & 34.1 & 34.7 & 33.3 \\
			\micon{gemini.png}~\mname{Gemma-3-27B-IT} & 57.3 & 57.6 & 51.6 & 65.0 & 45.3 & 47.8 & 43.5 & 55.8 & 63.0 & 63.4 & 57.1 & 57.7 & 56.4 \\
			\micon{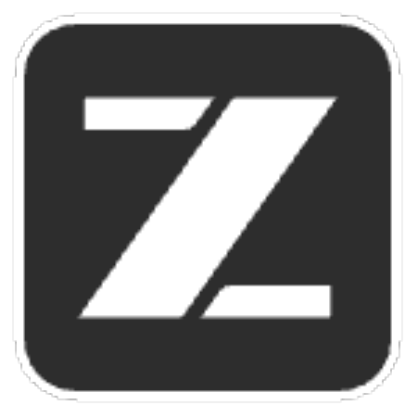}~\mname{GLM-4.6V} & 78.7 & 78.4 & 75.0 & 80.3 & 72.0 & 72.8 & 60.9 & 86.8 & 86.4 & 83.3 & 80.0 & 75.1 & \textbf{80.3} \\
			\micon{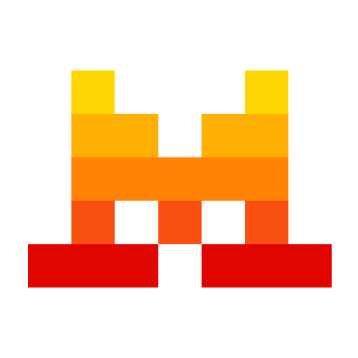}~\mname{Ministral-3-3B-Instruct} & 41.8 & 42.2 & 38.0 & 39.8 & 41.8 & 41.3 & 43.5 & 29.5 & 51.9 & 43.9 & 43.4 & 38.2 & 37.6 \\
			\micon{mistral.png}~\mname{Ministral-3-3B-Reasoning} & 45.2 & 42.7 & 42.5 & 47.6 & 44.7 & 51.4 & 37.7 & 48.1 & 46.9 & 47.7 & 46.5 & 42.2 & 45.3 \\
			\micon{mistral.png}~\mname{Ministral-3-14B-Instruct} & 56.5 & 54.6 & 51.4 & 65.6 & 54.4 & 56.2 & 47.8 & 50.4 & 69.1 & 59.9 & 57.8 & 53.8 & 50.4 \\
			\micon{mistral.png}~\mname{Ministral-3-14B-Reasoning} & 72.7 & 71.2 & 70.3 & 71.1 & 73.6 & 73.9 & 55.8 & 79.8 & 86.4 & 75.5 & 73.4 & 71.4 & 69.2 \\
			\midrule
			\multicolumn{14}{l}{\emph{Medical-specialized VLMs}} \\
			\midrule
			\micon{gemini.png}~\mname{MedGemma-1.5-4B-IT} & 44.4 & 44.3 & 39.9 & 49.3 & 36.5 & 43.8 & 50.0 & 55.0 & 48.1 & 45.6 & 48.1 & 36.1 & 36.8 \\
			\micon{gemini.png}~\mname{MedGemma-4B-IT} & 34.4 & 31.3 & 28.1 & 35.7 & 35.8 & 31.5 & 39.1 & 30.2 & 29.6 & 40.3 & 34.8 & 34.5 & 20.5 \\
			\micon{gemini.png}~\mname{MedGemma-27B-IT} & 56.8 & 57.9 & 44.4 & 63.9 & 50.0 & 54.7 & 41.3 & 62.8 & 58.0 & 62.0 & 56.3 & 59.2 & 42.7 \\
			\micon{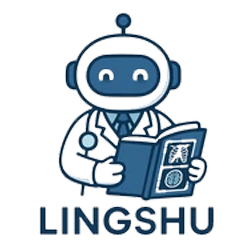}~\mname{Lingshu-7B} & 46.6 & 44.8 & 44.1 & 53.7 & 39.3 & 44.6 & 47.1 & 51.2 & 54.3 & 49.1 & 48.7 & 41.6 & 43.6 \\
			\micon{lingshu_big.png}~\mname{Lingshu-32B} & \textbf{66.1} & \textbf{65.5} & 60.2 & 74.1 & \textbf{64.2} & \textbf{62.0} & 54.3 & 77.5 & \textbf{81.5} & \textbf{68.0} & \textbf{66.8} & \textbf{64.4} & \textbf{65.0} \\
			\micon{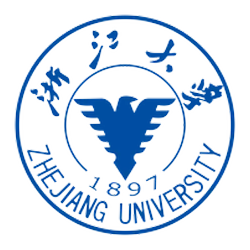}~\mname{Hulu-Med-7B} & 43.9 & 41.1 & 41.0 & 48.6 & 39.3 & 44.6 & 42.8 & 48.8 & 45.7 & 47.3 & 45.6 & 40.2 & 36.8 \\
			\micon{hulu_med_big.png}~\mname{Hulu-Med-32B} & 64.7 & 61.8 & \textbf{63.5} & \textbf{74.5} & 59.4 & 59.8 & \textbf{60.9} & \textbf{81.4} & 70.4 & 66.7 & 66.5 & 60.6 & 59.0 \\
			\midrule
			\multicolumn{14}{l}{\emph{Korean-specialized VLMs}} \\
			\midrule
			\micon{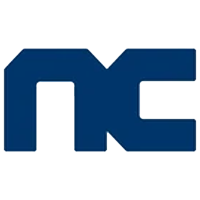}~\mname{VARCO-VISION-2.0-14B} & \textbf{43.2} & \textbf{42.4} & 35.4 & \textbf{46.3} & 39.6 & \textbf{41.3} & 32.6 & \textbf{48.1} & \textbf{44.4} & \textbf{48.5} & \textbf{43.7} & \textbf{43.0} & 33.3 \\
			\micon{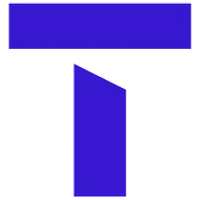}~\mname{A.X-4.0-VL-Light} & 41.8 & 40.5 & \textbf{37.0} & 43.5 & \textbf{43.4} & 35.5 & \textbf{39.9} & 35.7 & \textbf{44.4} & 46.3 & 43.1 & 38.7 & \textbf{41.9} \\
			\micon{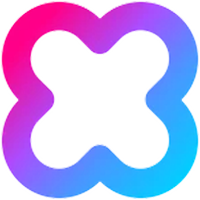}~\mname{HyperCLOVAX-SEED-Vision-Instruct-3B} & 26.3 & 25.7 & 21.5 & 21.8 & 27.0 & 23.9 & 26.1 & 26.4 & 13.6 & 31.0 & 27.4 & 24.1 & 21.4 \\
			\micon{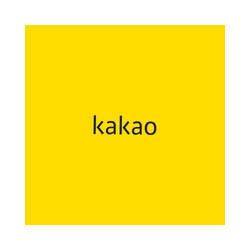}~\mname{Kanana-1.5-V-3B-Instruct} & 30.7 & 28.6 & 27.1 & 27.6 & 27.4 & 26.1 & 34.8 & 25.6 & 29.6 & 36.8 & 30.7 & 30.8 & 30.8 \\
			\bottomrule
		\end{tabular}%
	}
\end{table}

\textbf{Overall performance.} As shown in Table~\ref{tab:results_main_summary}, proprietary models dominate, far exceeding the majority baseline (22.4\%). Gemini-3.0-Pro and Gemini-3.0-Flash both achieve 96.9\%, followed by GPT-5 (93.9\%) and GPT-5-mini (90.1\%). Among open-source VLMs, Qwen3-VL-32B-Thinking leads at 83.7\%, followed by GLM-4.6V (78.7\%). Medical-specialized models trail notably---Lingshu-32B (66.1\%) and Hulu-Med-32B (64.7\%)---while Korean-specialized VLMs lag further: VARCO-VISION-2.0-14B (43.2\%) and A.X-4.0-VL-Light (41.8\%). Within each model family, performance scales consistently with parameter count (\textit{e.g.}, Qwen3-VL Instruct: 45.7\% at 4B $\rightarrow$ 76.5\% at 32B; InternVL3.5: 43.3\% at 4B $\rightarrow$ 67.2\% at 38B). Reasoning-oriented model variants further improve over instruction-tuned counterparts across all tested families, with gains of up to +20.4 percentage points (Qwen3-VL-8B) and +16.2 points (Ministral-3-14B).

\textbf{Modality-wise variation.} Performance varies substantially across imaging modalities. Using Gemini-3.0-Pro as reference, MRI and PBS (peripheral blood smear) achieve 100\% accuracy, followed by CT (97.9\%), Other (97.4\%), ultrasound (97.2\%), and X-ray (97.0\%), with ECG (95.9\%), endoscopy (93.5\%), and NST (91.3\%) at the lower end. Averaging across all 51 evaluated models reveals consistent modality-level difficulty patterns: NST, CT, and endoscopy remain the most challenging modalities, while Other, MRI, and ECG yield the highest accuracy. These modality-level gaps generally persist across model families, suggesting that modality-specific visual characteristics may contribute to performance variation.

\textbf{Multi-image reasoning.} Most models perform best on single-image questions and show degraded performance on multi-image items. Qwen3-VL-32B-Thinking scores 85.2\% on single-image items but drops to 80.5\% on two-image and 76.9\% on 3+ image questions. Even Gemini-3.0-Pro degrades from 97.7\% (single) to 87.2\% (3+ images). This trend is consistent across all models: the 51-model average drops from 57.0\% (1 image) to 53.8\% (2 images) to 50.3\% (3+ images), indicating that integrating evidence across multiple medical images remains an open challenge for current VLMs.

\textbf{Medical domain adaptation.} Medical domain adaptation yields results that depend on both model scale and training methodology. Since Lingshu is built on Qwen2.5-VL, we can directly measure the effect of medical fine-tuning: Lingshu-32B improves by +3.1 percentage points over Qwen2.5-VL-32B-Instruct (66.1\% vs.\ 63.0\%), while Lingshu-7B shows no gain over Qwen2.5-VL-7B-Instruct (both 46.6\%), suggesting that medical knowledge acquired at smaller scales may not fully generalize to a cross-lingual exam setting. MedGemma further highlights the role of training methodology: while MedGemma-4B shows negligible gain over base Gemma~3 (34.4\% vs.\ 34.3\%), the revised MedGemma-1.5-4B achieves 44.4\% (+10.1 percentage points over the same base), though at 27B scale MedGemma-27B (56.8\%) remains comparable to Gemma-3-27B (57.3\%). Despite these per-family improvements, an absolute gap persists: even the best medical-specialized model, Lingshu-32B (66.1\%), trails the best general-purpose model, Qwen3-VL-32B-Thinking (83.7\%), by 17.6 percentage points. These results indicate that backbone capacity remains a primary factor in overall performance, but domain-adaptation training---when properly designed---can provide meaningful gains; both the scale of the base model and the quality of the adaptation recipe matter for effective medical specialization.

\subsection{KorMedMCQA-Mixed Results}
\label{sec:combined_year_eval}
While KorMedMCQA-V focuses exclusively on multimodal items, actual medical licensing examinations administer a mixture of text-only and image-based questions.
To provide a more realistic evaluation that mirrors real exam conditions for vision-language models, we construct combined-year benchmarks (KorMedMCQA-Mixed) by integrating multimodal items from KorMedMCQA-V with text-only items from the KorMedMCQA doctor test split for corresponding exam years (2022--2023).
For concreteness, we report results on two representative years: KorMedMCQA-Mixed-2022 and KorMedMCQA-Mixed-2023.

We summarize performance using (i) Text accuracy on text-only items, (ii) Vision accuracy on multimodal items, and (iii) Total accuracy on the union of both item types (\textit{i.e.}, the fraction of correctly answered questions in the combined set).
KorMedMCQA-Mixed-2022 contains 134 text-only items and 147 multimodal items (total 281 items).
KorMedMCQA-Mixed-2023 contains 150 text-only items and 157 multimodal items (total 307 items).

\begin{table}[H]
	\centering
	\scriptsize
	\setlength{\tabcolsep}{3pt}
	\renewcommand{\arraystretch}{0.92}
	\caption{Summary of additional experimental results on the combined-year benchmarks (KorMedMCQA-Mixed-2022/2023, selected models). \textbf{Bold} indicates the best score within each column for each model category. The average row is computed over all 51 evaluated models. Parenthesized numbers below column headers indicate the number of questions in each split. Full results are available in Appendix~\ref{app:full_results_mixed}.}
	\label{tab:results_mixed_summary}
	\resizebox{0.8\linewidth}{!}{%
		\begin{tabular}{lcccccc}
			\toprule
			\textbf{Model}                                                     & \multicolumn{3}{c}{\textbf{KorMedMCQA-Mixed-2022}} & \multicolumn{3}{c}{\textbf{KorMedMCQA-Mixed-2023}}                                                                     \\
			\cmidrule(lr){2-4} \cmidrule(lr){5-7}
			                                                                   & \textbf{Text}                                      & \textbf{Vision}                                    & \textbf{Total} & \textbf{Text} & \textbf{Vision} & \textbf{Total} \\
			                                                                   & \scriptsize{(134)} & \scriptsize{(147)} & \scriptsize{(281)} & \scriptsize{(150)} & \scriptsize{(157)} & \scriptsize{(307)} \\
			\midrule
			\textit{Average (n=51)} & 59.0 & 55.0 & 56.9 & 59.5 & 57.7 & 58.6 \\
			\mname{Always choose majority label (E)} & 20.0 & 24.7 & 22.4 & 22.0 & 23.6 & 22.8 \\
			\midrule
			\multicolumn{7}{l}{\emph{Proprietary models}} \\
			\midrule
			\micon{gemini.png}~\mname{Gemini-3.0-Pro} & \textbf{98.5} & \textbf{97.3} & \textbf{97.9} & \textbf{99.3} & 94.3 & \textbf{96.7} \\
			\micon{gemini.png}~\mname{Gemini-3.0-Flash} & 97.0 & 95.2 & 96.1 & 98.0 & \textbf{94.9} & 96.4 \\
			\micon{openai.pdf}~\mname{GPT-5-2025-08-07} & 95.6 & 89.7 & 92.5 & 98.0 & 91.1 & 94.5 \\
			\micon{openai.pdf}~\mname{GPT-5-mini-2025-08-07} & 91.9 & 90.4 & 91.1 & 93.3 & 92.4 & 92.8 \\
			\midrule
			\multicolumn{7}{l}{\emph{General-purpose VLMs}} \\
			\midrule
			\micon{intervl.png}~\mname{InternVL3.5-4B} & 40.0 & 43.4 & 41.8 & 41.8 & 41.8 & 41.8 \\
			\micon{intervl.png}~\mname{InternVL3.5-8B} & 48.9 & 47.0 & 47.9 & 48.2 & 52.7 & 50.5 \\
			\micon{intervl.png}~\mname{InternVL3.5-38B} & 61.5 & 63.5 & 62.5 & 64.9 & 69.6 & 67.3 \\
			\micon{qwen.pdf}~\mname{Qwen2.5-VL-7B-Instruct} & 48.6 & 39.3 & 43.8 & 47.8 & 50.7 & 49.3 \\
			\micon{qwen.pdf}~\mname{Qwen2.5-VL-32B-Instruct} & 59.8 & 56.2 & 57.9 & 66.2 & 66.2 & 66.2 \\
			\micon{qwen.pdf}~\mname{Qwen3-VL-4B-Instruct} & 46.4 & 41.3 & 43.8 & 50.0 & 47.6 & 48.8 \\
			\micon{qwen.pdf}~\mname{Qwen3-VL-4B-Thinking} & 70.9 & 63.7 & 67.1 & 71.6 & 66.5 & 68.9 \\
			\micon{qwen.pdf}~\mname{Qwen3-VL-8B-Instruct} & 56.8 & 53.4 & 55.0 & 53.6 & 53.5 & 53.5 \\
			\micon{qwen.pdf}~\mname{Qwen3-VL-8B-Thinking} & 76.5 & 73.7 & 75.1 & 80.0 & 77.5 & 78.7 \\
			\micon{qwen.pdf}~\mname{Qwen3-VL-32B-Instruct} & 71.9 & 71.9 & 71.9 & 80.2 & 79.8 & 80.0 \\
			\micon{qwen.pdf}~\mname{Qwen3-VL-32B-Thinking} & \textbf{84.2} & \textbf{79.7} & \textbf{81.9} & \textbf{88.0} & \textbf{84.9} & \textbf{86.4} \\
			\micon{gemini.png}~\mname{Gemma-3-4B-IT} & 37.0 & 31.3 & 34.0 & 30.0 & 33.5 & 31.8 \\
			\micon{gemini.png}~\mname{Gemma-3-27B-IT} & 63.7 & 55.5 & 59.4 & 58.4 & 56.7 & 57.5 \\
			\micon{zai.pdf}~\mname{GLM-4.6V} & 83.0 & 75.8 & 79.2 & 85.8 & 84.3 & 85.0 \\
			\micon{mistral.png}~\mname{Ministral-3-3B-Instruct} & 44.0 & 42.9 & 43.4 & 46.2 & 41.4 & 43.8 \\
			\micon{mistral.png}~\mname{Ministral-3-3B-Reasoning} & 54.3 & 49.1 & 51.6 & 51.3 & 48.4 & 49.8 \\
			\micon{mistral.png}~\mname{Ministral-3-14B-Instruct} & 61.5 & 54.1 & 57.7 & 62.2 & 63.1 & 62.6 \\
			\micon{mistral.png}~\mname{Ministral-3-14B-Reasoning} & 75.3 & 74.9 & 75.1 & 73.1 & 76.2 & 74.7 \\
			\midrule
			\multicolumn{7}{l}{\emph{Medical-specialized VLMs}} \\
			\midrule
			\micon{gemini.png}~\mname{MedGemma-1.5-4B-IT} & 54.8 & 43.6 & 49.0 & 51.8 & 43.7 & 47.7 \\
			\micon{gemini.png}~\mname{MedGemma-4B-IT} & 42.7 & 37.7 & 40.1 & 41.3 & 33.8 & 37.5 \\
			\micon{gemini.png}~\mname{MedGemma-27B-IT} & \textbf{64.4} & 52.3 & 58.1 & 61.3 & 57.7 & 59.5 \\
			\micon{lingshu_big.png}~\mname{Lingshu-7B} & 47.9 & 42.7 & 45.2 & 53.6 & 49.5 & 51.5 \\
			\micon{lingshu_big.png}~\mname{Lingshu-32B} & 63.0 & 62.6 & \textbf{62.8} & 64.4 & 71.3 & 68.0 \\
			\micon{hulu_med_big.png}~\mname{Hulu-Med-7B} & 54.3 & 42.7 & 48.3 & 51.3 & 50.5 & 50.9 \\
			\micon{hulu_med_big.png}~\mname{Hulu-Med-32B} & 62.5 & \textbf{63.0} & \textbf{62.8} & \textbf{69.3} & \textbf{71.5} & \textbf{70.5} \\
			\midrule
			\multicolumn{7}{l}{\emph{Korean-specialized VLMs}} \\
			\midrule
			\micon{nc.png}~\mname{VARCO-VISION-2.0-14B} & \textbf{59.3} & \textbf{46.6} & \textbf{52.7} & \textbf{57.3} & \textbf{45.9} & \textbf{51.5} \\
			\micon{skt.png}~\mname{A.X-4.0-VL-Light} & 54.8 & 42.9 & 48.6 & 53.1 & 44.6 & 48.8 \\
			\micon{naver.png}~\mname{HyperCLOVAX-SEED-Vision-Instruct-3B} & 33.3 & 29.5 & 31.3 & 32.7 & 26.1 & 29.3 \\
			\micon{kakao.png}~\mname{Kanana-1.5-V-3B-Instruct} & 37.0 & 30.8 & 33.8 & 41.3 & 34.4 & 37.8 \\
			\bottomrule
		\end{tabular}%
	}
\end{table}

\textbf{Overall performance.} As shown in Table~\ref{tab:results_mixed_summary}, model rankings on the Mixed benchmarks are broadly consistent with the vision-only results (Table~\ref{tab:results_main_summary}): proprietary models lead, followed by general-purpose open-source VLMs, with medical- and Korean-specialized models trailing, confirming benchmark integrity. Text-only items are easier on average, with a mean text--vision gap of +4.0 percentage points on Mixed-2022 (59.0\% vs.\ 55.0\%) and +1.8 percentage points on Mixed-2023 (59.5\% vs.\ 57.7\%). Notably, the text--vision gap varies widely across models, ranging from $-$9.0 percentage points (Qwen3-VL-2B-Instruct) to +16.3 percentage points (Ministral-3-8B-Reasoning) on Mixed-2022, indicating that vision items expose qualitative differences in visual reasoning ability that text-only evaluation alone cannot capture.

\textbf{Text--vision gap across model categories.} The magnitude of the text--vision gap varies systematically across model categories. Proprietary and general-purpose VLMs show the smallest category-average gaps: +2.8 percentage points each on Mixed-2022, narrowing to +4.2 and +0.4 percentage points respectively on Mixed-2023. Medical-specialized models exhibit a larger average gap (+6.8 on Mixed-2022, +1.9 on Mixed-2023) with high within-category variance: Hulu-Med-32B achieves higher vision than text accuracy on both years ($-$0.5 and $-$2.2 percentage points), while MedGemma-27B shows a +12.1 percentage point text advantage on Mixed-2022 (64.4\% text vs.\ 52.3\% vision), indicating that the effect of medical domain adaptation on visual reasoning is highly model-dependent. Korean-specialized models show the largest and most consistent gaps, averaging +8.3 percentage points on both Mixed-2022 and Mixed-2023; VARCO-VISION-2.0-14B exhibits +12.7 and +11.4 percentage point text advantages (59.3\%/46.6\% and 57.3\%/45.9\%), identifying medical visual reasoning as the primary weakness of Korean-specialized VLMs. These results indicate that general-purpose VLMs achieve relatively balanced cross-modal performance, medical domain adaptation does not consistently improve visual reasoning despite isolated successes at larger scales, and Korean-language specialization does not extend to medical visual reasoning.

\textbf{Pass/fail analysis.} Applying official medical licensing exam pass/fail criteria ($\geq$40\% per exam session, $\geq$60\% overall) to KorMedMCQA-Mixed, only 15 of 51 models (29.4\%) pass in 2022 and 20 (39.2\%) in 2023. All proprietary models pass both years, while no Korean-specialized model passes either year. Among open-source models, only reasoning-oriented or large-scale variants (\textit{e.g.}, Qwen3-VL-32B-Thinking, GLM-4.6V) consistently meet the threshold, and session-level failures---particularly on Session~1A (20 items)---frequently disqualify models whose overall accuracy would otherwise suffice (see Appendix~\ref{app:pass_fail} for full details).

\section{Conclusion}
We introduced KorMedMCQA-V, a multimodal multiple-choice question answering benchmark drawn from Korean Medical Licensing Examinations (2012--2023), comprising 1,534 questions with 2,043 images.
Together with the text-only KorMedMCQA benchmark, it forms a unified evaluation suite for Korean medical reasoning under both text-only and multimodal conditions.
Our zero-shot evaluation of over 50 VLMs across proprietary and open-source categories---spanning general-purpose, medical-specialized, and Korean-specialized families---yields three main insights:
(i) model scale and explicit reasoning capability are the dominant drivers of performance, outweighing both domain-specific fine-tuning and Korean language specialization;
(ii) multi-image reasoning remains a consistent bottleneck across model families; and
(iii) performance varies substantially across clinical imaging modalities, highlighting modality-specific gaps that aggregate accuracy alone does not capture.
We release the dataset and evaluation code to support reproducible research on Korean multimodal medical reasoning.

\subsection*{Limitations}
First, KorMedMCQA-V is limited to Korean physician licensing exams; generalization to other medical specialties, countries, or languages requires further study.
Second, the dataset spans 2012--2023; ongoing updates with more recent exam years would maintain relevance.
Third, we evaluate only zero-shot performance; few-shot learning and fine-tuning effects remain unexplored.
Fourth, some modalities have limited samples (MRI: 2.0\%, PBS: 2.4\%, NST: 2.6\%), constraining detailed per-modality analysis.

\subsection*{Reproducibility and Code Availability}
To support reproducible research, we publicly release the KorMedMCQA-V dataset via Hugging Face Datasets (\href{https://huggingface.co/datasets/seongsubae/KorMedMCQA-V}{https://huggingface.co/datasets/seongsubae/KorMedMCQA-V}) and the complete evaluation framework on GitHub (\href{https://github.com/baeseongsu/kormedmcqa_v}{https://github.com/baeseongsu/kormedmcqa\_v}).

\clearpage
\bibliographystyle{plain}
\bibliography{reference}

\clearpage
\appendix

\section*{Appendix: Table of Contents}
\begin{itemize}[leftmargin=1.5em, itemsep=2pt, topsep=0pt]
  \item[\ref{app:modality_prompt}.] \nameref{app:modality_prompt} \dotfill \pageref{app:modality_prompt}
  \item[\ref{app:models}.] \nameref{app:models} \dotfill \pageref{app:models}
  \item[\ref{app:full_results}.] \nameref{app:full_results} \dotfill \pageref{app:full_results}
  \item[\ref{app:full_results_mixed}.] \nameref{app:full_results_mixed} \dotfill \pageref{app:full_results_mixed}
  \item[\ref{app:pass_fail}.] \nameref{app:pass_fail} \dotfill \pageref{app:pass_fail}
\end{itemize}

\clearpage
\section{Prompt Template for Modality Annotation}
\label{app:modality_prompt}

We use a shared instruction template for model-based modality annotation; only the model endpoint changes across annotator models.
Model outputs are lowercased and matched to the canonical set (XRAY, CT, US, MRI, ECG, ENDOSCOPY, PBS, NST, OTHER); disagreements are sent to manual review. The image is provided as a separate vision input.

\begin{figure}[ht]
\begin{tcolorbox}[
   colback=white,
   colframe=black,
   fonttitle=\bfseries,
   coltitle=white,
   fontupper = \ttfamily\scriptsize,
   title = {Prompt Template for Modality Annotation},
   top=3pt, bottom=3pt, left=4pt, right=4pt,
]
\begin{Verbatim}[breaklines=true, fontsize=\scriptsize, breaksymbol={}]
# Role
You are a medical image classification expert. Your goal is to identify the type
of the provided image using both the image itself and the accompanying text.

# Task
Classify the given image into one of these 9 categories:
[Xray, CT, US, MRI, ECG, endoscopy, PBS, NST, other]

# Context Analysis Rules (Priority)
1. **Target Identification**: If the text explicitly introduces the image,
   prioritize those sentences, such as:
   - "~사진이다" (This is a photo of...)
   - "~소견이다" (This is a finding of...)
   - "~결과이다" (This is a result of...)
   - "다음은 ~이다" (The following is...)

2. **Handle Multiple Mentions**: If the text mentions multiple tests, prioritize
   the one linked to the visual presentation. Ignore tests mentioned only as
   part of the patient's history if they are not the source of the current image.

3. **Keyword Mapping (Korean to English)**:
   - "X선", "단순촬영", "방사선" -> Xray
   - "컴퓨터단층촬영", "CT" -> CT
   - "초음파" -> US
   - "자기공명영상", "MRI" -> MRI
   - "심전도" -> ECG
   - "내시경" -> endoscopy
   - "말초혈액도말" -> PBS
   - "비수축검사", "태동검사" -> NST
   - If the image shows physical skin findings ("피부 소견"), clinical photos,
     or anything else -> other

# Constraints
- Respond with exactly ONE word from the category list.
- Do not include any other text, explanation, or punctuation.
- Use lowercase.

# Input
- Text: {question_text}
- Image: (attached as an image input)

# Output
[category]
\end{Verbatim}
\end{tcolorbox}
\caption{Instruction template used for model-based modality annotation. The \texttt{\{question\_text\}} placeholder is replaced with the original Korean exam question text at inference time.}
\label{fig:modality_prompt}
\end{figure}

\clearpage
\section{List of Evaluated Models}
\label{app:models}

\begin{table}[H]
	\centering
	\small
	\setlength{\tabcolsep}{4pt}
	\renewcommand{\arraystretch}{0.95}
	\caption{VLM baselines used in our experiments. ``Engine'' indicates the inference framework used for evaluation: API (proprietary endpoints), vLLM (server-based), or HF (HuggingFace Transformers with custom handlers where applicable).}
	\label{tab:vlm_baselines}
	\resizebox{0.78\linewidth}{!}{%
		\begin{tabular}{lll}
			\toprule
			\textbf{Model}                      & \textbf{Source}                                       & \textbf{Engine} \\
			\midrule
			\multicolumn{3}{l}{\emph{Proprietary models (n=5)}}                                                                                        \\
			\midrule
			\micon{openai.pdf}~GPT-5.2                              & OpenAI API                                            & API        \\
			\micon{openai.pdf}~GPT-5-mini-2025-08-07               & OpenAI API                                            & API        \\
			\micon{openai.pdf}~GPT-5-2025-08-07                    & OpenAI API                                            & API        \\
			\micon{gemini.png}~Gemini-3.0-Pro                      & Google API                                            & API        \\
			\micon{gemini.png}~Gemini-3.0-Flash                    & Google API                                            & API        \\
			\midrule
			\multicolumn{3}{l}{\emph{General-purpose VLMs (n=32)}}                                                                               \\
			\midrule
			\micon{intervl.png}~InternVL3\_5-1B                     & OpenGVLab/InternVL3\_5-1B                             & vLLM        \\
			\micon{intervl.png}~InternVL3\_5-2B                     & OpenGVLab/InternVL3\_5-2B                             & vLLM        \\
			\micon{intervl.png}~InternVL3\_5-4B                     & OpenGVLab/InternVL3\_5-4B                             & vLLM        \\
			\micon{intervl.png}~InternVL3\_5-8B                     & OpenGVLab/InternVL3\_5-8B                             & vLLM        \\
			\micon{intervl.png}~InternVL3\_5-14B                    & OpenGVLab/InternVL3\_5-14B                            & vLLM        \\
			\micon{intervl.png}~InternVL3\_5-30B-A3B                & OpenGVLab/InternVL3\_5-30B-A3B                        & vLLM        \\
			\micon{intervl.png}~InternVL3\_5-38B                    & OpenGVLab/InternVL3\_5-38B                            & vLLM        \\
			\micon{qwen.pdf}~Qwen2.5-VL-7B-Instruct              & Qwen/Qwen2.5-VL-7B-Instruct                           & vLLM        \\
			\micon{qwen.pdf}~Qwen2.5-VL-32B-Instruct             & Qwen/Qwen2.5-VL-32B-Instruct                          & vLLM        \\
			\micon{qwen.pdf}~Qwen3-VL-2B-Instruct                & Qwen/Qwen3-VL-2B-Instruct                             & vLLM        \\
			\micon{qwen.pdf}~Qwen3-VL-2B-Thinking                & Qwen/Qwen3-VL-2B-Thinking                             & vLLM        \\
			\micon{qwen.pdf}~Qwen3-VL-4B-Instruct                & Qwen/Qwen3-VL-4B-Instruct                             & vLLM        \\
			\micon{qwen.pdf}~Qwen3-VL-4B-Thinking                & Qwen/Qwen3-VL-4B-Thinking                             & vLLM        \\
			\micon{qwen.pdf}~Qwen3-VL-8B-Instruct                & Qwen/Qwen3-VL-8B-Instruct                             & vLLM        \\
			\micon{qwen.pdf}~Qwen3-VL-8B-Thinking                & Qwen/Qwen3-VL-8B-Thinking                             & vLLM        \\
			\micon{qwen.pdf}~Qwen3-VL-30B-A3B-Instruct           & Qwen/Qwen3-VL-30B-A3B-Instruct                        & vLLM        \\
			\micon{qwen.pdf}~Qwen3-VL-30B-A3B-Thinking           & Qwen/Qwen3-VL-30B-A3B-Thinking                        & vLLM        \\
			\micon{qwen.pdf}~Qwen3-VL-32B-Instruct               & Qwen/Qwen3-VL-32B-Instruct                            & vLLM        \\
			\micon{qwen.pdf}~Qwen3-VL-32B-Thinking               & Qwen/Qwen3-VL-32B-Thinking                            & vLLM        \\
			\micon{gemini.png}~Gemma-3-4B-IT                       & google/gemma-3-4b-it                                  & vLLM        \\
			\micon{gemini.png}~Gemma-3-12B-IT                      & google/gemma-3-12b-it                                 & vLLM        \\
			\micon{gemini.png}~Gemma-3-27B-IT                      & google/gemma-3-27b-it                                 & vLLM        \\
			\micon{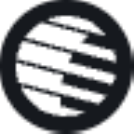}~Kimi-VL-A3B-Instruct                & moonshotai/Kimi-VL-A3B-Instruct                       & vLLM        \\
			\micon{moonshot.pdf}~Kimi-VL-A3B-Thinking                & moonshotai/Kimi-VL-A3B-Thinking                       & vLLM        \\
			\micon{zai.pdf}~GLM-4.6V                            & zai-org/GLM-4.6V                                      & vLLM        \\
			\micon{zai.pdf}~GLM-4.6V-Flash                      & zai-org/GLM-4.6V-Flash                                & vLLM        \\
			\micon{mistral.png}~Ministral-3-3B-Instruct-2512-BF16   & mistralai/Ministral-3-3B-Instruct-2512-BF16           & vLLM        \\
			\micon{mistral.png}~Ministral-3-3B-Reasoning-2512       & mistralai/Ministral-3-3B-Reasoning-2512               & vLLM        \\
			\micon{mistral.png}~Ministral-3-8B-Instruct-2512-BF16   & mistralai/Ministral-3-8B-Instruct-2512-BF16           & vLLM        \\
			\micon{mistral.png}~Ministral-3-8B-Reasoning-2512       & mistralai/Ministral-3-8B-Reasoning-2512               & vLLM        \\
			\micon{mistral.png}~Ministral-3-14B-Instruct-2512-BF16  & mistralai/Ministral-3-14B-Instruct-2512-BF16          & vLLM        \\
			\micon{mistral.png}~Ministral-3-14B-Reasoning-2512      & mistralai/Ministral-3-14B-Reasoning-2512              & vLLM        \\
			\midrule
			\multicolumn{3}{l}{\emph{Medical-specialized VLMs (n=9)}}                                                                                  \\
			\midrule
			\micon{gemini.png}~MedGemma-1.5-4B-IT                  & google/medgemma-1.5-4b-it                             & vLLM        \\
			\micon{gemini.png}~MedGemma-4B-IT                      & google/medgemma-4b-it                                 & vLLM        \\
			\micon{gemini.png}~MedGemma-27B-IT                     & google/medgemma-27b-it                                & vLLM        \\
			\micon{lingshu_big.png}~Lingshu-7B                          & lingshu-medical-mllm/Lingshu-7B                       & vLLM        \\
			\micon{lingshu_big.png}~Lingshu-32B                         & lingshu-medical-mllm/Lingshu-32B                      & vLLM        \\
			\micon{hulu_med_big.png}~Hulu-Med-4B                         & ZJU-AI4H/Hulu-Med-4B                                  & HF        \\
			\micon{hulu_med_big.png}~Hulu-Med-7B                         & ZJU-AI4H/Hulu-Med-7B                                  & HF        \\
			\micon{hulu_med_big.png}~Hulu-Med-14B                        & ZJU-AI4H/Hulu-Med-14B                                 & HF        \\
			\micon{hulu_med_big.png}~Hulu-Med-32B                        & ZJU-AI4H/Hulu-Med-32B                                 & HF        \\
			\midrule
			\multicolumn{3}{l}{\emph{Korean-specialized VLMs (n=5)}}                                                                                   \\
			\midrule
			\micon{nc.png}~VARCO-VISION-2.0-1.7B               & NCSOFT/VARCO-VISION-2.0-1.7B                          & vLLM        \\
			\micon{nc.png}~VARCO-VISION-2.0-14B                & NCSOFT/VARCO-VISION-2.0-14B                           & vLLM        \\
			\micon{skt.png}~A.X-4.0-VL-Light                    & skt/A.X-4.0-VL-Light                                  & HF        \\
			\micon{naver.png}~HyperCLOVAX-SEED-Vision-Instruct-3B & naver-hyperclovax/HyperCLOVAX-SEED-Vision-Instruct-3B & HF        \\
			\micon{kakao.png}~Kanana-1.5-V-3B-Instruct            & kakaocorp/kanana-1.5-v-3b-instruct                    & HF        \\
			\bottomrule
		\end{tabular}%
	}
\end{table}

\clearpage
\section{Detailed KorMedMCQA-V Results}
\label{app:full_results}

\begin{table}[H]
	\centering
	\scriptsize
	\setlength{\tabcolsep}{3pt}
	\renewcommand{\arraystretch}{0.92}
	\caption{Main experimental results on KorMedMCQA-V. We report mean ± std across three random seeds (42, 43, 44). \textbf{Bold} indicates the best mean score within each column for each model category. Parenthesized numbers below column headers indicate the number of image instances for modality columns and the number of questions for image count columns.}
	\label{tab:results_main}
	\resizebox{\linewidth}{!}{%
		\begin{tabular}{lccccccccccccc}
			\toprule
			\textbf{Model}                                                     & \textbf{Overall} & \multicolumn{9}{c}{\textbf{Modality}} & \multicolumn{3}{c}{\textbf{\# Images}}                                                                                                                                                     \\
			\cmidrule(lr){3-11} \cmidrule(lr){12-14}
			                                                                   &                  & \textbf{XRAY}                         & \textbf{CT}                            & \textbf{ECG} & \textbf{US} & \textbf{Endo.} & \textbf{NST} & \textbf{PBS} & \textbf{MRI} & \textbf{Other} & \textbf{1} & \textbf{2} & \textbf{3+} \\
			                                                                   &                  & \scriptsize{(586)} & \scriptsize{(336)} & \scriptsize{(164)} & \scriptsize{(138)} & \scriptsize{(122)} & \scriptsize{(54)} & \scriptsize{(49)} & \scriptsize{(40)} & \scriptsize{(554)} & \scriptsize{(1,069)} & \scriptsize{(426)} & \scriptsize{(39)} \\
			\midrule
			\textit{Average (n=51)} & 55.9 & 55.2 & 51.5 & 58.5 & 54.5 & 52.9 & 50.5 & 57.7 & 59.0 & 59.2 & 57.0 & 53.8 & 50.3 \\
			\mname{Always choose majority label (E)} & 22.4 & 23.5 & 21.3 & 12.2 & 25.5 & 19.6 & 30.4 & 23.3 & 25.9 & 22.7 & 23.2 & 20.4 & 23.1 \\
			\midrule
			\multicolumn{14}{l}{\emph{Proprietary VLMs (n=5)}} \\
			\midrule
			\micon{gemini.png}~\mname{gemini-3.0-pro} & \textbf{$96.9$} & \textbf{$97.0$} & \textbf{$97.9$} & \textbf{$95.9$} & $97.2$ & $93.5$ & $91.3$ & \textbf{$100.0$} & \textbf{$100.0$} & $97.4$ & \textbf{$97.7$} & \textbf{$96.0$} & $87.2$ \\
			\micon{gemini.png}~\mname{gemini-3.0-flash} & \textbf{$96.9$} & $96.2$ & \textbf{$97.9$} & $93.9$ & $96.2$ & \textbf{$94.6$} & \textbf{$93.5$} & \textbf{$100.0$} & \textbf{$100.0$} & \textbf{$98.3$} & \textbf{$97.6$} & $95.5$ & \textbf{$92.3$} \\
			\micon{openai.pdf}~\mname{gpt-5.2-2025-12-11} & $93.9$ & $94.2$ & $89.6$ & $92.9$ & \textbf{$98.1$} & $90.2$ & $84.8$ & \textbf{$100.0$} & $96.3$ & $95.5$ & $94.7$ & $92.7$ & $84.6$ \\
			\micon{openai.pdf}~\mname{gpt-5-2025-08-07} & $93.9$ & $93.6$ & $89.1$ & $89.8$ & $97.2$ & $92.4$ & $87.0$ & \textbf{$100.0$} & \textbf{$100.0$} & $96.3$ & $94.7$ & $92.0$ & \textbf{$92.3$} \\
			\micon{openai.pdf}~\mname{gpt-5-mini-2025-08-07} & $90.1$ & $89.1$ & $85.4$ & $85.7$ & $90.6$ & $88.0$ & $89.1$ & $97.7$ & $96.3$ & $93.3$ & $91.3$ & $87.3$ & $87.2$ \\
			\midrule
			\multicolumn{14}{l}{\emph{General open-source VLMs (n=32)}} \\
			\midrule
			
			\micon{intervl.png}~\mname{InternVL3.5-1B} & $22.4 \pm 0.4$ & $21.4 \pm 0.3$ & $22.6 \pm 0.8$ & $20.1 \pm 0.6$ & $19.5 \pm 6.8$ & $18.8 \pm 2.3$ & $30.4 \pm 0.0$ & $18.6 \pm 2.3$ & $6.2 \pm 2.1$ & $25.6 \pm 1.7$ & $22.6 \pm 0.4$ & $22.6 \pm 2.1$ & $13.7 \pm 3.0$ \\
			\micon{intervl.png}~\mname{InternVL3.5-2B} & $29.8 \pm 0.9$ & $29.8 \pm 1.7$ & $27.3 \pm 2.2$ & $32.0 \pm 1.6$ & $28.6 \pm 6.3$ & $23.6 \pm 1.7$ & $32.6 \pm 2.2$ & $28.7 \pm 4.8$ & $24.7 \pm 5.7$ & $32.1 \pm 1.1$ & $30.3 \pm 1.2$ & $28.4 \pm 0.6$ & $31.6 \pm 5.3$ \\
			\micon{intervl.png}~\mname{InternVL3.5-4B} & $43.3 \pm 0.8$ & $43.9 \pm 0.9$ & $41.0 \pm 1.3$ & $44.9 \pm 2.7$ & $43.7 \pm 2.0$ & $40.6 \pm 3.5$ & $38.4 \pm 3.3$ & $47.3 \pm 2.7$ & $42.0 \pm 4.3$ & $43.9 \pm 2.4$ & $44.2 \pm 0.7$ & $41.6 \pm 1.6$ & $35.9 \pm 2.6$ \\
			\micon{intervl.png}~\mname{InternVL3.5-8B} & $47.4 \pm 1.5$ & $44.8 \pm 1.4$ & $44.3 \pm 1.5$ & $53.6 \pm 2.2$ & $49.1 \pm 1.3$ & $41.8 \pm 0.8$ & $42.4 \pm 10.8$ & $45.3 \pm 1.6$ & $61.1 \pm 2.6$ & $50.6 \pm 2.4$ & $49.0 \pm 1.3$ & $44.7 \pm 1.8$ & $32.1 \pm 1.8$ \\
			\micon{intervl.png}~\mname{InternVL3.5-14B} & $53.6 \pm 0.5$ & $54.2 \pm 0.6$ & $48.4 \pm 1.0$ & $58.8 \pm 2.4$ & $49.4 \pm 1.4$ & $48.9 \pm 1.1$ & $52.9 \pm 1.3$ & $53.5 \pm 2.3$ & $56.8 \pm 2.1$ & $55.8 \pm 0.8$ & $53.6 \pm 0.6$ & $53.7 \pm 0.9$ & $53.0 \pm 1.5$ \\
			\micon{intervl.png}~\mname{InternVL3.5-30B-A3B} & $66.4 \pm 0.5$ & $68.9 \pm 1.5$ & $60.1 \pm 1.1$ & $71.1 \pm 1.2$ & $65.4 \pm 2.7$ & $56.5 \pm 1.1$ & $54.3 \pm 5.8$ & $75.2 \pm 1.3$ & $75.3 \pm 2.1$ & $67.5 \pm 0.1$ & $67.3 \pm 0.5$ & $64.5 \pm 1.7$ & $62.4 \pm 3.0$ \\
			\micon{intervl.png}~\mname{InternVL3.5-38B} & $67.2 \pm 0.6$ & $66.7 \pm 0.6$ & $62.2 \pm 1.6$ & $72.8 \pm 1.6$ & $61.9 \pm 1.4$ & $64.1 \pm 2.9$ & $50.7 \pm 9.8$ & $76.0 \pm 3.6$ & $79.0 \pm 2.1$ & $70.6 \pm 0.4$ & $68.6 \pm 0.4$ & $64.2 \pm 1.6$ & $61.5 \pm 5.1$ \\
			\micon{qwen.pdf}~\mname{Qwen2.5-VL-7B-Instruct} & $46.6 \pm 0.1$ & $46.9 \pm 0.3$ & $42.7 \pm 0.0$ & $50.0 \pm 0.0$ & $41.5 \pm 0.0$ & $42.8 \pm 0.6$ & $43.5 \pm 0.0$ & $39.5 \pm 2.3$ & $44.4 \pm 0.0$ & $50.1 \pm 0.3$ & $46.8 \pm 0.1$ & $45.4 \pm 0.1$ & $52.1 \pm 1.5$ \\
			\micon{qwen.pdf}~\mname{Qwen2.5-VL-32B-Instruct} & $63.0 \pm 0.1$ & $62.3 \pm 0.5$ & $56.9 \pm 0.3$ & $71.8 \pm 0.6$ & $64.5 \pm 1.1$ & $59.1 \pm 0.6$ & $52.2 \pm 0.0$ & $67.4 \pm 0.0$ & $70.4 \pm 0.0$ & $65.2 \pm 0.2$ & $64.0 \pm 0.0$ & $61.4 \pm 0.4$ & $54.7 \pm 1.5$ \\
			\micon{qwen.pdf}~\mname{Qwen3-VL-2B-Instruct} & $31.3 \pm 0.2$ & $33.0 \pm 0.9$ & $29.9 \pm 0.6$ & $30.3 \pm 2.1$ & $29.6 \pm 2.9$ & $34.4 \pm 1.3$ & $34.8 \pm 0.0$ & $17.1 \pm 3.6$ & $27.2 \pm 2.1$ & $31.5 \pm 0.8$ & $31.3 \pm 0.5$ & $31.1 \pm 0.4$ & $34.2 \pm 1.5$ \\
			\micon{qwen.pdf}~\mname{Qwen3-VL-2B-Thinking} & $40.4 \pm 0.6$ & $40.8 \pm 1.5$ & $36.8 \pm 2.7$ & $43.9 \pm 1.0$ & $41.2 \pm 3.6$ & $43.5 \pm 0.0$ & $29.7 \pm 5.5$ & $38.8 \pm 1.3$ & $35.8 \pm 4.3$ & $41.3 \pm 0.7$ & $41.9 \pm 0.9$ & $37.5 \pm 1.3$ & $29.9 \pm 7.8$ \\
			\micon{qwen.pdf}~\mname{Qwen3-VL-4B-Instruct} & $45.7 \pm 0.5$ & $41.7 \pm 0.9$ & $42.0 \pm 0.3$ & $53.1 \pm 1.0$ & $41.2 \pm 1.4$ & $47.5 \pm 0.6$ & $46.4 \pm 5.0$ & $46.5 \pm 2.3$ & $35.8 \pm 2.1$ & $50.9 \pm 0.6$ & $47.1 \pm 0.2$ & $43.3 \pm 1.3$ & $32.5 \pm 1.5$ \\
			\micon{qwen.pdf}~\mname{Qwen3-VL-4B-Thinking} & $65.9 \pm 0.3$ & $63.1 \pm 0.8$ & $64.1 \pm 1.8$ & $70.4 \pm 4.1$ & $68.2 \pm 1.1$ & $66.3 \pm 0.0$ & $46.4 \pm 3.3$ & $82.2 \pm 3.6$ & $60.5 \pm 5.7$ & $68.7 \pm 0.6$ & $68.1 \pm 0.5$ & $61.5 \pm 0.8$ & $54.7 \pm 1.5$ \\
			\micon{qwen.pdf}~\mname{Qwen3-VL-8B-Instruct} & $53.8 \pm 0.5$ & $50.4 \pm 0.4$ & $46.5 \pm 1.5$ & $56.8 \pm 2.1$ & $55.3 \pm 2.0$ & $51.1 \pm 0.0$ & $50.7 \pm 3.3$ & $46.5 \pm 2.3$ & $55.6 \pm 0.0$ & $60.8 \pm 0.9$ & $55.9 \pm 0.4$ & $49.1 \pm 1.4$ & $47.9 \pm 1.5$ \\
			\micon{qwen.pdf}~\mname{Qwen3-VL-8B-Thinking} & $74.2 \pm 0.0$ & $75.1 \pm 0.2$ & $67.9 \pm 2.4$ & $74.8 \pm 2.1$ & $76.7 \pm 3.0$ & $69.6 \pm 4.3$ & $68.8 \pm 5.5$ & $84.5 \pm 1.3$ & $72.8 \pm 5.7$ & $75.8 \pm 1.5$ & $75.7 \pm 0.3$ & $71.4 \pm 0.8$ & $65.8 \pm 1.5$ \\
			\micon{qwen.pdf}~\mname{Qwen3-VL-30B-A3B-Instruct} & $71.7 \pm 1.0$ & $72.2 \pm 1.1$ & $63.7 \pm 1.3$ & $75.5 \pm 1.8$ & $70.1 \pm 3.0$ & $65.9 \pm 2.5$ & $63.0 \pm 2.2$ & $72.9 \pm 4.8$ & $76.5 \pm 2.1$ & $75.5 \pm 1.1$ & $72.3 \pm 0.9$ & $71.0 \pm 1.2$ & $62.4 \pm 1.5$ \\
			\micon{qwen.pdf}~\mname{Qwen3-VL-30B-A3B-Thinking} & $80.4 \pm 1.4$ & $80.7 \pm 2.5$ & $73.8 \pm 1.5$ & $82.3 \pm 0.6$ & \textbf{$81.1 \pm 0.9$} & $75.4 \pm 3.5$ & $70.3 \pm 3.3$ & $90.7 \pm 0.0$ & $87.7 \pm 2.1$ & $83.0 \pm 1.4$ & $81.7 \pm 1.3$ & $78.3 \pm 1.7$ & $67.5 \pm 3.9$ \\
			\micon{qwen.pdf}~\mname{Qwen3-VL-32B-Instruct} & $76.5 \pm 0.2$ & $75.9 \pm 0.7$ & $71.9 \pm 1.4$ & $83.3 \pm 1.6$ & $77.0 \pm 1.4$ & $67.8 \pm 0.6$ & $64.5 \pm 1.3$ & $86.0 \pm 2.3$ & $85.2 \pm 3.7$ & $79.1 \pm 0.7$ & $77.9 \pm 0.1$ & $74.5 \pm 0.5$ & $61.5 \pm 2.6$ \\
			\micon{qwen.pdf}~\mname{Qwen3-VL-32B-Thinking} & \textbf{$83.7 \pm 0.2$} & \textbf{$83.9 \pm 1.0$} & \textbf{$78.0 \pm 1.8$} & \textbf{$86.4 \pm 1.2$} & \textbf{$81.1 \pm 1.6$} & \textbf{$76.4 \pm 1.7$} & \textbf{$71.7 \pm 2.2$} & \textbf{$95.3 \pm 0.0$} & \textbf{$88.9 \pm 3.7$} & \textbf{$87.0 \pm 1.2$} & \textbf{$85.2 \pm 0.4$} & \textbf{$80.5 \pm 0.7$} & $76.9 \pm 2.6$ \\
			\micon{gemini.png}~\mname{Gemma-3-4B-IT} & $34.3 \pm 0.1$ & $32.7 \pm 0.0$ & $27.6 \pm 0.0$ & $36.7 \pm 0.0$ & $39.6 \pm 0.0$ & $26.1 \pm 0.0$ & $30.4 \pm 0.0$ & $27.9 \pm 0.0$ & $48.1 \pm 0.0$ & $38.7 \pm 0.2$ & $34.1 \pm 0.1$ & $34.7 \pm 0.0$ & $33.3 \pm 0.0$ \\
			\micon{gemini.png}~\mname{Gemma-3-12B-IT} & $54.3 \pm 0.0$ & $51.6 \pm 0.1$ & $53.5 \pm 0.3$ & $62.6 \pm 0.6$ & $50.9 \pm 0.0$ & $51.1 \pm 0.0$ & $41.3 \pm 0.0$ & $46.5 \pm 0.0$ & $48.1 \pm 0.0$ & $59.6 \pm 0.1$ & $55.4 \pm 0.1$ & $52.5 \pm 0.1$ & $45.3 \pm 1.5$ \\
			\micon{gemini.png}~\mname{Gemma-3-27B-IT} & $57.3 \pm 0.1$ & $57.6 \pm 0.1$ & $51.6 \pm 0.0$ & $65.0 \pm 0.6$ & $45.3 \pm 0.0$ & $47.8 \pm 0.0$ & $43.5 \pm 0.0$ & $55.8 \pm 0.0$ & $63.0 \pm 0.0$ & $63.4 \pm 0.0$ & $57.1 \pm 0.1$ & $57.7 \pm 0.0$ & $56.4 \pm 0.0$ \\
			\micon{moonshot.pdf}~\mname{Kimi-VL-A3B-Instruct} & $30.8 \pm 0.9$ & $30.3 \pm 2.0$ & $28.8 \pm 1.6$ & $22.8 \pm 2.6$ & $31.8 \pm 4.4$ & $29.0 \pm 1.3$ & $37.0 \pm 2.2$ & $20.2 \pm 2.7$ & $28.4 \pm 2.1$ & $34.5 \pm 0.3$ & $32.3 \pm 1.1$ & $26.3 \pm 0.2$ & $40.2 \pm 1.5$ \\
			\micon{moonshot.pdf}~\mname{Kimi-VL-A3B-Thinking} & $37.6 \pm 0.3$ & $37.2 \pm 0.3$ & $28.6 \pm 0.7$ & $43.4 \pm 0.7$ & $34.4 \pm 0.7$ & $34.2 \pm 0.8$ & $31.5 \pm 4.6$ & $38.4 \pm 8.2$ & $44.4 \pm 10.5$ & $42.2 \pm 0.6$ & $39.9 \pm 1.1$ & $32.4 \pm 0.7$ & $32.1 \pm 9.1$ \\
			\micon{zai.pdf}~\mname{GLM-4.6V} & $78.7 \pm 0.2$ & $78.4 \pm 1.2$ & $75.0 \pm 1.4$ & $80.3 \pm 1.6$ & $72.0 \pm 1.4$ & $72.8 \pm 1.1$ & $60.9 \pm 2.2$ & $86.8 \pm 2.7$ & $86.4 \pm 2.1$ & $83.3 \pm 0.9$ & $80.0 \pm 0.5$ & $75.1 \pm 1.6$ & \textbf{$80.3 \pm 1.5$} \\
			\micon{zai.pdf}~\mname{GLM-4.6V-Flash} & $65.5 \pm 0.2$ & $63.3 \pm 1.2$ & $59.5 \pm 0.8$ & $68.4 \pm 1.0$ & $60.4 \pm 0.9$ & $62.7 \pm 3.8$ & $48.6 \pm 5.0$ & $65.1 \pm 4.7$ & $80.2 \pm 2.1$ & $72.2 \pm 1.8$ & $66.3 \pm 0.2$ & $63.6 \pm 0.5$ & $63.2 \pm 3.9$ \\
			\micon{mistral.png}~\mname{Ministral-3-3B-Instruct-2512} & $41.8 \pm 0.7$ & $42.2 \pm 1.3$ & $38.0 \pm 1.0$ & $39.8 \pm 1.0$ & $41.8 \pm 0.5$ & $41.3 \pm 1.1$ & $43.5 \pm 3.8$ & $29.5 \pm 3.6$ & $51.9 \pm 3.7$ & $43.9 \pm 0.6$ & $43.4 \pm 0.7$ & $38.2 \pm 0.5$ & $37.6 \pm 3.9$ \\
			\micon{mistral.png}~\mname{Ministral-3-3B-Reasoning-2512} & $45.2 \pm 0.6$ & $42.7 \pm 1.5$ & $42.5 \pm 2.1$ & $47.6 \pm 1.6$ & $44.7 \pm 1.4$ & $51.4 \pm 1.7$ & $37.7 \pm 3.3$ & $48.1 \pm 8.2$ & $46.9 \pm 2.1$ & $47.7 \pm 0.6$ & $46.5 \pm 0.6$ & $42.2 \pm 1.5$ & $45.3 \pm 1.5$ \\
			\micon{mistral.png}~\mname{Ministral-3-8B-Instruct-2512} & $53.8 \pm 0.7$ & $55.1 \pm 0.7$ & $43.9 \pm 0.8$ & $53.4 \pm 1.2$ & $58.5 \pm 4.1$ & $51.4 \pm 1.7$ & $47.1 \pm 1.3$ & $52.7 \pm 1.3$ & $56.8 \pm 4.3$ & $56.7 \pm 0.9$ & $55.5 \pm 0.8$ & $50.4 \pm 0.5$ & $44.4 \pm 1.5$ \\
			\micon{mistral.png}~\mname{Ministral-3-8B-Reasoning-2512} & $65.1 \pm 0.7$ & $64.0 \pm 1.3$ & $62.0 \pm 5.0$ & $66.7 \pm 3.6$ & $66.4 \pm 4.4$ & $59.8 \pm 6.1$ & $57.2 \pm 3.3$ & $68.2 \pm 5.9$ & $71.6 \pm 4.3$ & $68.0 \pm 0.8$ & $66.2 \pm 1.2$ & $62.9 \pm 0.5$ & $59.0 \pm 6.8$ \\
			\micon{mistral.png}~\mname{Ministral-3-14B-Instruct-2512} & $56.5 \pm 0.4$ & $54.6 \pm 0.4$ & $51.4 \pm 0.8$ & $65.6 \pm 1.6$ & $54.4 \pm 2.4$ & $56.2 \pm 3.1$ & $47.8 \pm 2.2$ & $50.4 \pm 1.3$ & $69.1 \pm 2.1$ & $59.9 \pm 0.5$ & $57.8 \pm 0.1$ & $53.8 \pm 0.9$ & $50.4 \pm 3.9$ \\
			\micon{mistral.png}~\mname{Ministral-3-14B-Reasoning-2512} & $72.7 \pm 1.1$ & $71.2 \pm 2.0$ & $70.3 \pm 1.8$ & $71.1 \pm 3.1$ & $73.6 \pm 3.4$ & $73.9 \pm 2.2$ & $55.8 \pm 4.5$ & $79.8 \pm 7.5$ & $86.4 \pm 2.1$ & $75.5 \pm 0.7$ & $73.4 \pm 1.8$ & $71.4 \pm 0.7$ & $69.2 \pm 4.4$ \\
			\midrule
			\multicolumn{14}{l}{\emph{Medical-specific VLMs (n=9)}} \\
			\midrule
			\micon{gemini.png}~\mname{MedGemma-1.5-4B-IT} & $44.4 \pm 1.1$ & $44.3 \pm 0.5$ & $39.9 \pm 0.3$ & $49.3 \pm 2.6$ & $36.5 \pm 3.9$ & $43.8 \pm 1.3$ & $50.0 \pm 7.5$ & $55.0 \pm 3.6$ & $48.1 \pm 3.7$ & $45.6 \pm 1.4$ & $48.1 \pm 1.4$ & $36.1 \pm 1.5$ & $36.8 \pm 3.0$ \\
			\micon{gemini.png}~\mname{MedGemma-4B-IT} & $34.4 \pm 0.1$ & $31.3 \pm 0.2$ & $28.1 \pm 0.0$ & $35.7 \pm 0.0$ & $35.8 \pm 0.0$ & $31.5 \pm 0.0$ & $39.1 \pm 0.0$ & $30.2 \pm 0.0$ & $29.6 \pm 0.0$ & $40.3 \pm 0.0$ & $34.8 \pm 0.1$ & $34.5 \pm 0.2$ & $20.5 \pm 0.0$ \\
			\micon{gemini.png}~\mname{MedGemma-27B-IT} & $56.8 \pm 0.1$ & $57.9 \pm 0.2$ & $44.4 \pm 0.3$ & $63.9 \pm 0.6$ & $50.0 \pm 0.0$ & $54.7 \pm 0.6$ & $41.3 \pm 0.0$ & $62.8 \pm 0.0$ & $58.0 \pm 2.1$ & $62.0 \pm 0.1$ & $56.3 \pm 0.2$ & $59.2 \pm 0.2$ & $42.7 \pm 1.5$ \\
			\micon{lingshu_big.png}~\mname{Lingshu-7B} & $46.6 \pm 0.1$ & $44.8 \pm 0.1$ & $44.1 \pm 0.3$ & $53.7 \pm 0.6$ & $39.3 \pm 0.5$ & $44.6 \pm 0.0$ & $47.1 \pm 1.3$ & $51.2 \pm 0.0$ & $54.3 \pm 2.1$ & $49.1 \pm 0.2$ & $48.7 \pm 0.1$ & $41.6 \pm 0.1$ & $43.6 \pm 0.0$ \\
			\micon{lingshu_big.png}~\mname{Lingshu-32B} & \textbf{$66.1 \pm 0.1$} & \textbf{$65.5 \pm 0.2$} & $60.2 \pm 0.8$ & $74.1 \pm 0.6$ & \textbf{$64.2 \pm 0.0$} & \textbf{$62.0 \pm 0.0$} & $54.3 \pm 0.0$ & $77.5 \pm 1.3$ & \textbf{$81.5 \pm 0.0$} & \textbf{$68.0 \pm 0.2$} & \textbf{$66.8 \pm 0.1$} & \textbf{$64.4 \pm 0.4$} & \textbf{$65.0 \pm 1.5$} \\
			\micon{hulu_med_big.png}~\mname{Hulu-Med-4B} & $46.1 \pm 0.2$ & $45.4 \pm 1.4$ & $39.2 \pm 1.7$ & $48.6 \pm 2.1$ & $45.6 \pm 1.1$ & $44.2 \pm 2.3$ & $46.4 \pm 3.3$ & $48.1 \pm 3.6$ & $54.3 \pm 2.1$ & $49.0 \pm 0.5$ & $47.2 \pm 0.2$ & $44.3 \pm 0.8$ & $35.0 \pm 3.9$ \\
			\micon{hulu_med_big.png}~\mname{Hulu-Med-7B} & $43.9 \pm 1.0$ & $41.1 \pm 1.6$ & $41.0 \pm 0.8$ & $48.6 \pm 5.0$ & $39.3 \pm 2.0$ & $44.6 \pm 1.1$ & $42.8 \pm 6.6$ & $48.8 \pm 4.7$ & $45.7 \pm 4.3$ & $47.3 \pm 1.1$ & $45.6 \pm 1.1$ & $40.2 \pm 1.0$ & $36.8 \pm 3.0$ \\
			\micon{hulu_med_big.png}~\mname{Hulu-Med-14B} & $55.1 \pm 0.3$ & $54.3 \pm 0.0$ & $50.7 \pm 1.7$ & $55.8 \pm 6.5$ & $52.8 \pm 3.8$ & $53.3 \pm 3.9$ & $59.4 \pm 3.3$ & $54.3 \pm 1.3$ & $53.1 \pm 5.7$ & $58.2 \pm 1.3$ & $55.8 \pm 0.6$ & $53.6 \pm 0.5$ & $51.3 \pm 5.1$ \\
			\micon{hulu_med_big.png}~\mname{Hulu-Med-32B} & $64.7 \pm 0.0$ & $61.8 \pm 0.0$ & \textbf{$63.5 \pm 0.0$} & \textbf{$74.5 \pm 0.0$} & $59.4 \pm 0.0$ & $59.8 \pm 0.0$ & \textbf{$60.9 \pm 0.0$} & \textbf{$81.4 \pm 0.0$} & $70.4 \pm 0.0$ & $66.7 \pm 0.0$ & $66.5 \pm 0.0$ & $60.6 \pm 0.0$ & $59.0 \pm 0.0$ \\
			\midrule
			\multicolumn{14}{l}{\emph{Korean-specific VLMs (n=5)}} \\
			\midrule
			\micon{nc.png}~\mname{VARCO-VISION-2.0-1.7B} & $24.4 \pm 0.0$ & $24.6 \pm 0.0$ & $21.9 \pm 0.0$ & $24.5 \pm 0.0$ & $21.7 \pm 0.0$ & $20.7 \pm 0.0$ & $28.3 \pm 0.0$ & $18.6 \pm 0.0$ & $22.2 \pm 0.0$ & $26.8 \pm 0.0$ & $24.9 \pm 0.0$ & $23.7 \pm 0.0$ & $17.9 \pm 0.0$ \\
			\micon{nc.png}~\mname{VARCO-VISION-2.0-14B} & \textbf{$43.2 \pm 0.1$} & \textbf{$42.4 \pm 0.1$} & $35.4 \pm 0.0$ & \textbf{$46.3 \pm 0.6$} & $39.6 \pm 0.0$ & \textbf{$41.3 \pm 0.0$} & $32.6 \pm 0.0$ & \textbf{$48.1 \pm 1.3$} & \textbf{$44.4 \pm 0.0$} & \textbf{$48.5 \pm 0.0$} & \textbf{$43.7 \pm 0.1$} & \textbf{$43.0 \pm 0.0$} & $33.3 \pm 0.0$ \\
			\micon{skt.png}~\mname{A.X-4.0-VL-Light} & $41.8 \pm 1.3$ & $40.5 \pm 0.7$ & \textbf{$37.0 \pm 2.8$} & $43.5 \pm 2.6$ & \textbf{$43.4 \pm 4.3$} & $35.5 \pm 1.3$ & \textbf{$39.9 \pm 4.5$} & $35.7 \pm 1.3$ & \textbf{$44.4 \pm 7.4$} & $46.3 \pm 2.1$ & $43.1 \pm 1.5$ & $38.7 \pm 1.2$ & \textbf{$41.9 \pm 1.5$} \\
			\micon{naver.png}~\mname{HyperCLOVAX-SEED-Vision-Instruct-3B} & $26.3 \pm 0.0$ & $25.7 \pm 0.5$ & $21.5 \pm 0.3$ & $21.8 \pm 0.6$ & $27.0 \pm 0.5$ & $23.9 \pm 1.1$ & $26.1 \pm 0.0$ & $26.4 \pm 1.3$ & $13.6 \pm 2.1$ & $31.0 \pm 0.4$ & $27.4 \pm 0.1$ & $24.1 \pm 0.6$ & $21.4 \pm 3.0$ \\
			\micon{kakao.png}~\mname{kanana-1.5-v-3b-instruct} & $30.7 \pm 0.0$ & $28.6 \pm 0.0$ & $27.1 \pm 0.0$ & $27.6 \pm 0.0$ & $27.4 \pm 0.0$ & $26.1 \pm 0.0$ & $34.8 \pm 0.0$ & $25.6 \pm 0.0$ & $29.6 \pm 0.0$ & $36.8 \pm 0.0$ & $30.7 \pm 0.0$ & $30.8 \pm 0.0$ & $30.8 \pm 0.0$ \\
			\bottomrule
		\end{tabular}%
	}
\end{table}

\clearpage
\section{Detailed KorMedMCQA-Mixed Results}
\label{app:full_results_mixed}

\begin{table}[H]
	\centering
	\scriptsize
	\setlength{\tabcolsep}{3pt}
	\renewcommand{\arraystretch}{0.92}
	\caption{Additional experimental results on the combined-year benchmarks (KorMedMCQA-Mixed-2022/2023). We report mean ± std across three random seeds (42, 43, 44). \textbf{Bold} indicates the best mean score within each column for each model category. Parenthesized numbers below column headers indicate the number of questions in each split.}
	\label{tab:results_additional}
	\resizebox{0.80\linewidth}{!}{%
		\begin{tabular}{lcccccc}
			\toprule
			\textbf{Model}                                                     & \multicolumn{3}{c}{\textbf{KorMedMCQA-Mixed-2022}} & \multicolumn{3}{c}{\textbf{KorMedMCQA-Mixed-2023}}                                                                     \\
			\cmidrule(lr){2-4} \cmidrule(lr){5-7}
			                                                                   & \textbf{Text}                                      & \textbf{Vision}                                    & \textbf{Total} & \textbf{Text} & \textbf{Vision} & \textbf{Total} \\
			                                                                   & \scriptsize{(134)} & \scriptsize{(147)} & \scriptsize{(281)} & \scriptsize{(150)} & \scriptsize{(157)} & \scriptsize{(307)} \\
			\midrule
			\textit{Average (n=51)} & 59.0 & 55.0 & 56.9 & 59.5 & 57.7 & 58.6 \\
			\mname{Always choose majority label (E)} & 20.0 & 24.7 & 22.4 & 22.0 & 23.6 & 22.8 \\
			\midrule
			\multicolumn{7}{l}{\emph{Proprietary VLMs (n=5)}} \\
			\midrule
			\micon{gemini.png}~\mname{gemini-3.0-pro} & \textbf{$98.5$} & \textbf{$97.3$} & \textbf{$97.9$} & \textbf{$99.3$} & $94.3$ & \textbf{$96.7$} \\
			\micon{gemini.png}~\mname{gemini-3.0-flash} & $97.0$ & $95.2$ & $96.1$ & $98.0$ & \textbf{$94.9$} & $96.4$ \\
			\micon{openai.pdf}~\mname{gpt-5.2-2025-12-11} & $95.6$ & $91.8$ & $93.6$ & $97.3$ & $92.4$ & $94.8$ \\
			\micon{openai.pdf}~\mname{gpt-5-2025-08-07} & $95.6$ & $89.7$ & $92.5$ & $98.0$ & $91.1$ & $94.5$ \\
			\micon{openai.pdf}~\mname{gpt-5-mini-2025-08-07} & $91.9$ & $90.4$ & $91.1$ & $93.3$ & $92.4$ & $92.8$ \\
			\midrule
			\multicolumn{7}{l}{\emph{General open-source VLMs (n=32)}} \\
			\midrule
			
			\micon{intervl.png}~\mname{InternVL3.5-1B} & $19.8 \pm 0.4$ & $26.3 \pm 2.1$ & $23.1 \pm 0.9$ & $20.7 \pm 0.7$ & $21.7 \pm 2.3$ & $21.2 \pm 0.9$ \\
			\micon{intervl.png}~\mname{InternVL3.5-2B} & $29.4 \pm 2.6$ & $30.4 \pm 1.4$ & $29.9 \pm 0.9$ & $32.7 \pm 3.3$ & $30.1 \pm 2.2$ & $31.4 \pm 0.5$ \\
			\micon{intervl.png}~\mname{InternVL3.5-4B} & $40.0 \pm 1.5$ & $43.4 \pm 2.1$ & $41.8 \pm 1.1$ & $41.8 \pm 1.4$ & $41.8 \pm 3.8$ & $41.8 \pm 1.6$ \\
			\micon{intervl.png}~\mname{InternVL3.5-8B} & $48.9 \pm 3.8$ & $47.0 \pm 2.8$ & $47.9 \pm 2.9$ & $48.2 \pm 2.7$ & $52.7 \pm 5.3$ & $50.5 \pm 4.0$ \\
			\micon{intervl.png}~\mname{InternVL3.5-14B} & $49.9 \pm 0.9$ & $55.5 \pm 0.7$ & $52.8 \pm 0.2$ & $58.7 \pm 2.3$ & $57.5 \pm 0.4$ & $58.1 \pm 1.0$ \\
			\micon{intervl.png}~\mname{InternVL3.5-30B-A3B} & $59.8 \pm 1.1$ & $62.3 \pm 2.4$ & $61.1 \pm 0.7$ & $65.6 \pm 1.7$ & $67.7 \pm 1.9$ & $66.7 \pm 1.8$ \\
			\micon{intervl.png}~\mname{InternVL3.5-38B} & $61.5 \pm 2.0$ & $63.5 \pm 0.4$ & $62.5 \pm 0.7$ & $64.9 \pm 3.8$ & $69.6 \pm 1.6$ & $67.3 \pm 2.3$ \\
			\micon{qwen.pdf}~\mname{Qwen2.5-VL-7B-Instruct} & $48.6 \pm 0.4$ & $39.3 \pm 1.0$ & $43.8 \pm 0.7$ & $47.8 \pm 0.4$ & $50.7 \pm 0.4$ & $49.3 \pm 0.2$ \\
			\micon{qwen.pdf}~\mname{Qwen2.5-VL-32B-Instruct} & $59.8 \pm 0.4$ & $56.2 \pm 0.0$ & $57.9 \pm 0.2$ & $66.2 \pm 0.4$ & $66.2 \pm 0.0$ & $66.2 \pm 0.2$ \\
			\micon{qwen.pdf}~\mname{Qwen3-VL-2B-Instruct} & $28.4 \pm 0.9$ & $37.4 \pm 1.0$ & $33.1 \pm 0.6$ & $32.0 \pm 1.2$ & $30.4 \pm 1.0$ & $31.2 \pm 0.7$ \\
			\micon{qwen.pdf}~\mname{Qwen3-VL-2B-Thinking} & $50.4 \pm 2.6$ & $38.6 \pm 3.1$ & $44.2 \pm 0.9$ & $44.7 \pm 1.3$ & $42.7 \pm 0.6$ & $43.6 \pm 0.6$ \\
			\micon{qwen.pdf}~\mname{Qwen3-VL-4B-Instruct} & $46.4 \pm 0.9$ & $41.3 \pm 1.0$ & $43.8 \pm 0.6$ & $50.0 \pm 2.0$ & $47.6 \pm 0.7$ & $48.8 \pm 1.3$ \\
			\micon{qwen.pdf}~\mname{Qwen3-VL-4B-Thinking} & $70.9 \pm 2.3$ & $63.7 \pm 3.0$ & $67.1 \pm 2.6$ & $71.6 \pm 2.0$ & $66.5 \pm 2.0$ & $68.9 \pm 1.7$ \\
			\micon{qwen.pdf}~\mname{Qwen3-VL-8B-Instruct} & $56.8 \pm 2.3$ & $53.4 \pm 0.7$ & $55.0 \pm 0.9$ & $53.6 \pm 1.0$ & $53.5 \pm 0.0$ & $53.5 \pm 0.5$ \\
			\micon{qwen.pdf}~\mname{Qwen3-VL-8B-Thinking} & $76.5 \pm 2.6$ & $73.7 \pm 1.6$ & $75.1 \pm 2.0$ & $80.0 \pm 2.0$ & $77.5 \pm 0.7$ & $78.7 \pm 1.3$ \\
			\micon{qwen.pdf}~\mname{Qwen3-VL-30B-A3B-Instruct} & $67.9 \pm 1.5$ & $68.9 \pm 0.4$ & $68.4 \pm 0.9$ & $68.9 \pm 1.0$ & $74.1 \pm 2.6$ & $71.6 \pm 1.0$ \\
			\micon{qwen.pdf}~\mname{Qwen3-VL-30B-A3B-Thinking} & $80.7 \pm 2.7$ & $79.5 \pm 1.2$ & $80.1 \pm 1.8$ & $84.2 \pm 0.4$ & $81.5 \pm 0.0$ & $82.8 \pm 0.2$ \\
			\micon{qwen.pdf}~\mname{Qwen3-VL-32B-Instruct} & $71.9 \pm 1.3$ & $71.9 \pm 1.4$ & $71.9 \pm 1.3$ & $80.2 \pm 1.0$ & $79.8 \pm 1.0$ & $80.0 \pm 0.2$ \\
			\micon{qwen.pdf}~\mname{Qwen3-VL-32B-Thinking} & \textbf{$84.2 \pm 1.9$} & \textbf{$79.7 \pm 1.4$} & \textbf{$81.9 \pm 1.6$} & \textbf{$88.0 \pm 1.2$} & \textbf{$84.9 \pm 1.3$} & \textbf{$86.4 \pm 1.0$} \\
			\micon{gemini.png}~\mname{Gemma-3-4B-IT} & $37.0 \pm 0.0$ & $31.3 \pm 0.4$ & $34.0 \pm 0.2$ & $30.0 \pm 0.0$ & $33.5 \pm 0.4$ & $31.8 \pm 0.2$ \\
			\micon{gemini.png}~\mname{Gemma-3-12B-IT} & $63.0 \pm 0.0$ & $55.9 \pm 0.4$ & $59.3 \pm 0.2$ & $57.3 \pm 0.0$ & $61.8 \pm 0.0$ & $59.6 \pm 0.0$ \\
			\micon{gemini.png}~\mname{Gemma-3-27B-IT} & $63.7 \pm 0.0$ & $55.5 \pm 0.0$ & $59.4 \pm 0.0$ & $58.4 \pm 0.4$ & $56.7 \pm 0.0$ & $57.5 \pm 0.2$ \\
			\micon{moonshot.pdf}~\mname{Kimi-VL-A3B-Instruct} & $27.7 \pm 2.3$ & $31.7 \pm 1.4$ & $29.8 \pm 1.8$ & $29.3 \pm 1.2$ & $26.5 \pm 3.0$ & $27.9 \pm 2.1$ \\
			\micon{moonshot.pdf}~\mname{Kimi-VL-A3B-Thinking} & $42.2 \pm 1.0$ & $40.8 \pm 1.5$ & $41.5 \pm 1.3$ & $38.0 \pm 0.9$ & $35.7 \pm 0.0$ & $36.8 \pm 0.5$ \\
			\micon{zai.pdf}~\mname{GLM-4.6V} & $83.0 \pm 2.0$ & $75.8 \pm 1.7$ & $79.2 \pm 1.2$ & $85.8 \pm 1.7$ & $84.3 \pm 1.0$ & $85.0 \pm 1.2$ \\
			\micon{zai.pdf}~\mname{GLM-4.6V-Flash} & $68.4 \pm 0.9$ & $65.5 \pm 1.6$ & $66.9 \pm 0.7$ & $66.2 \pm 1.4$ & $65.8 \pm 1.9$ & $66.0 \pm 1.6$ \\
			\micon{mistral.png}~\mname{Ministral-3-3B-Instruct-2512} & $44.0 \pm 1.1$ & $42.9 \pm 4.0$ & $43.4 \pm 2.5$ & $46.2 \pm 1.7$ & $41.4 \pm 1.9$ & $43.8 \pm 1.0$ \\
			\micon{mistral.png}~\mname{Ministral-3-3B-Reasoning-2512} & $54.3 \pm 2.3$ & $49.1 \pm 2.8$ & $51.6 \pm 1.6$ & $51.3 \pm 1.2$ & $48.4 \pm 1.3$ & $49.8 \pm 0.9$ \\
			\micon{mistral.png}~\mname{Ministral-3-8B-Instruct-2512} & $58.3 \pm 0.4$ & $48.2 \pm 0.8$ & $53.0 \pm 0.6$ & $55.6 \pm 0.4$ & $59.4 \pm 3.2$ & $57.5 \pm 1.7$ \\
			\micon{mistral.png}~\mname{Ministral-3-8B-Reasoning-2512} & $74.3 \pm 0.9$ & $58.0 \pm 3.4$ & $65.8 \pm 1.6$ & $72.2 \pm 1.0$ & $64.5 \pm 3.9$ & $68.3 \pm 1.8$ \\
			\micon{mistral.png}~\mname{Ministral-3-14B-Instruct-2512} & $61.5 \pm 0.7$ & $54.1 \pm 1.8$ & $57.7 \pm 1.2$ & $62.2 \pm 1.0$ & $63.1 \pm 1.9$ & $62.6 \pm 1.5$ \\
			\micon{mistral.png}~\mname{Ministral-3-14B-Reasoning-2512} & $75.3 \pm 0.9$ & $74.9 \pm 4.2$ & $75.1 \pm 1.9$ & $73.1 \pm 3.3$ & $76.2 \pm 2.6$ & $74.7 \pm 2.7$ \\
			\midrule
			\multicolumn{7}{l}{\emph{Medical-specific VLMs (n=9)}} \\
			\midrule
			\micon{gemini.png}~\mname{MedGemma-1.5-4B-IT} & $54.8 \pm 1.3$ & $43.6 \pm 1.4$ & $49.0 \pm 0.8$ & $51.8 \pm 2.1$ & $43.7 \pm 0.4$ & $47.7 \pm 1.0$ \\
			\micon{gemini.png}~\mname{MedGemma-4B-IT} & $42.7 \pm 0.4$ & $37.7 \pm 0.0$ & $40.1 \pm 0.2$ & $41.3 \pm 0.0$ & $33.8 \pm 0.0$ & $37.5 \pm 0.0$ \\
			\micon{gemini.png}~\mname{MedGemma-27B-IT} & \textbf{$64.4 \pm 0.0$} & $52.3 \pm 0.4$ & $58.1 \pm 0.2$ & $61.3 \pm 0.0$ & $57.7 \pm 0.4$ & $59.5 \pm 0.2$ \\
			\micon{lingshu_big.png}~\mname{Lingshu-7B} & $47.9 \pm 0.4$ & $42.7 \pm 0.8$ & $45.2 \pm 0.4$ & $53.6 \pm 0.4$ & $49.5 \pm 0.7$ & $51.5 \pm 0.3$ \\
			\micon{lingshu_big.png}~\mname{Lingshu-32B} & $63.0 \pm 0.0$ & $62.6 \pm 0.4$ & \textbf{$62.8 \pm 0.2$} & $64.4 \pm 0.4$ & $71.3 \pm 0.0$ & $68.0 \pm 0.2$ \\
			\micon{hulu_med_big.png}~\mname{Hulu-Med-4B} & $54.1 \pm 2.7$ & $45.4 \pm 1.7$ & $49.6 \pm 2.1$ & $46.7 \pm 2.3$ & $49.0 \pm 1.1$ & $47.9 \pm 1.5$ \\
			\micon{hulu_med_big.png}~\mname{Hulu-Med-7B} & $54.3 \pm 2.1$ & $42.7 \pm 0.4$ & $48.3 \pm 0.8$ & $51.3 \pm 2.3$ & $50.5 \pm 1.3$ & $50.9 \pm 1.5$ \\
			\micon{hulu_med_big.png}~\mname{Hulu-Med-14B} & $62.7 \pm 2.6$ & $55.3 \pm 2.8$ & $58.8 \pm 2.4$ & $65.1 \pm 2.5$ & $60.9 \pm 1.0$ & $63.0 \pm 1.5$ \\
			\micon{hulu_med_big.png}~\mname{Hulu-Med-32B} & $62.5 \pm 0.4$ & \textbf{$63.0 \pm 2.4$} & \textbf{$62.8 \pm 1.1$} & \textbf{$69.3 \pm 2.0$} & \textbf{$71.5 \pm 1.9$} & \textbf{$70.5 \pm 1.8$} \\
			\midrule
			\multicolumn{7}{l}{\emph{Korean-specific VLMs (n=5)}} \\
			\midrule
			\micon{nc.png}~\mname{VARCO-VISION-2.0-1.7B} & $35.6 \pm 0.0$ & $28.8 \pm 0.0$ & $32.0 \pm 0.0$ & $34.7 \pm 0.0$ & $26.8 \pm 0.0$ & $30.6 \pm 0.0$ \\
			\micon{nc.png}~\mname{VARCO-VISION-2.0-14B} & \textbf{$59.3 \pm 0.0$} & \textbf{$46.6 \pm 0.0$} & \textbf{$52.7 \pm 0.0$} & \textbf{$57.3 \pm 0.0$} & \textbf{$45.9 \pm 0.0$} & \textbf{$51.5 \pm 0.0$} \\
			\micon{skt.png}~\mname{A.X-4.0-VL-Light} & $54.8 \pm 2.6$ & $42.9 \pm 1.4$ & $48.6 \pm 1.6$ & $53.1 \pm 2.3$ & $44.6 \pm 4.5$ & $48.8 \pm 1.1$ \\
			\micon{naver.png}~\mname{HyperCLOVAX-SEED-Vision-Instruct-3B} & $33.3 \pm 0.0$ & $29.5 \pm 0.7$ & $31.3 \pm 0.4$ & $32.7 \pm 0.0$ & $26.1 \pm 1.3$ & $29.3 \pm 0.7$ \\
			\micon{kakao.png}~\mname{kanana-1.5-v-3b-instruct} & $37.0 \pm 0.0$ & $30.8 \pm 0.0$ & $33.8 \pm 0.0$ & $41.3 \pm 0.0$ & $34.4 \pm 0.0$ & $37.8 \pm 0.0$ \\
			\bottomrule
		\end{tabular}%
	}
\end{table}

\clearpage
\section{Pass/Fail Analysis on KorMedMCQA-Mixed}
\label{app:pass_fail}

We apply the official medical licensing exam pass/fail criteria to each model's predictions on KorMedMCQA-Mixed.
The 2022--2023 exam requires $\geq$40\% in each exam session (Session~1A: Q1--20; Session~1B: Q21--80; Sessions~2--4 combined) and $\geq$60\% overall.
A model passes only if all conditions are met simultaneously.
We evaluate on the available subset and apply percentage-based thresholds proportionally, as some items were excluded (R-type removal, image availability).

\begin{table}[H]
	\centering
	\fontsize{6.5pt}{7.5pt}\selectfont
	\setlength{\tabcolsep}{2pt}
	\renewcommand{\arraystretch}{0.85}
	\caption{KMLE pass/fail analysis based on official exam criteria ($\geq$40\% per section, $\geq$60\% overall). \colorbox{red!12}{Red-shaded} cells fall below the threshold. Open-source models: mean $\pm$ std across three seeds (42, 43, 44).}
	\label{tab:pass_fail}
	\resizebox{\linewidth}{!}{%
		\begin{tabular}{l ccccc ccccc}
			\toprule
			\textbf{Model} & \multicolumn{5}{c}{\textbf{2022}} & \multicolumn{5}{c}{\textbf{2023}} \\
			\cmidrule(lr){2-6} \cmidrule(lr){7-11}
			 & \textbf{S1-A} & \textbf{S1-B} & \textbf{S2--4} & \textbf{Total} & \textbf{Pass}
			 & \textbf{S1-A} & \textbf{S1-B} & \textbf{S2--4} & \textbf{Total} & \textbf{Pass} \\
			 & \scriptsize{(20)} & \scriptsize{(59)} & \scriptsize{(202)} & \scriptsize{(281)} &  & \scriptsize{(20)} & \scriptsize{(59)} & \scriptsize{(228)} & \scriptsize{(307)} &  \\
			\midrule
			\multicolumn{11}{l}{\emph{Proprietary VLMs (n=5; pass: 5/5 in 2022, 5/5 in 2023)}} \\
			\midrule
			\micon{gemini.png}~\mname{gemini-3.0-pro} & $95.0 \pm 0.0$ & $98.3 \pm 0.0$ & $98.0 \pm 0.0$ & $97.9 \pm 0.0$ & \cellcolor{green!15}\cmark & $95.0 \pm 0.0$ & $98.3 \pm 0.0$ & $96.5 \pm 0.0$ & $96.7 \pm 0.0$ & \cellcolor{green!15}\cmark \\
			\micon{gemini.png}~\mname{gemini-3.0-flash} & $90.0 \pm 0.0$ & $96.6 \pm 0.0$ & $96.5 \pm 0.0$ & $96.1 \pm 0.0$ & \cellcolor{green!15}\cmark & $85.0 \pm 0.0$ & $98.3 \pm 0.0$ & $96.9 \pm 0.0$ & $96.4 \pm 0.0$ & \cellcolor{green!15}\cmark \\
			\micon{openai.pdf}~\mname{gpt-5.2-2025-12-11} & $80.0 \pm 0.0$ & $93.2 \pm 0.0$ & $95.0 \pm 0.0$ & $93.6 \pm 0.0$ & \cellcolor{green!15}\cmark & $90.0 \pm 0.0$ & $94.9 \pm 0.0$ & $95.2 \pm 0.0$ & $94.8 \pm 0.0$ & \cellcolor{green!15}\cmark \\
			\micon{openai.pdf}~\mname{gpt-5-2025-08-07} & $80.0 \pm 0.0$ & $98.3 \pm 0.0$ & $92.1 \pm 0.0$ & $92.5 \pm 0.0$ & \cellcolor{green!15}\cmark & $85.0 \pm 0.0$ & $96.6 \pm 0.0$ & $94.7 \pm 0.0$ & $94.5 \pm 0.0$ & \cellcolor{green!15}\cmark \\
			\micon{openai.pdf}~\mname{gpt-5-mini-2025-08-07} & $65.0 \pm 0.0$ & $94.9 \pm 0.0$ & $92.6 \pm 0.0$ & $91.1 \pm 0.0$ & \cellcolor{green!15}\cmark & $70.0 \pm 0.0$ & $91.5 \pm 0.0$ & $95.2 \pm 0.0$ & $92.8 \pm 0.0$ & \cellcolor{green!15}\cmark \\
			\midrule
			\multicolumn{11}{l}{\emph{General open-source VLMs (n=32; pass: 9/32 in 2022, 12/32 in 2023)}} \\
			\midrule
			\micon{intervl.png}~\mname{InternVL3.5-1B} & \cellcolor{red!12}$13.3 \pm 2.9$ & \cellcolor{red!12}$24.3 \pm 1.0$ & \cellcolor{red!12}$23.8 \pm 1.0$ & \cellcolor{red!12}$23.1 \pm 0.9$ & \cellcolor{red!15}\xmark & \cellcolor{red!12}$18.3 \pm 5.8$ & \cellcolor{red!12}$17.5 \pm 1.0$ & \cellcolor{red!12}$22.4 \pm 1.5$ & \cellcolor{red!12}$21.2 \pm 0.9$ & \cellcolor{red!15}\xmark \\
			\micon{intervl.png}~\mname{InternVL3.5-2B} & \cellcolor{red!12}$33.3 \pm 2.9$ & \cellcolor{red!12}$35.6 \pm 1.7$ & \cellcolor{red!12}$27.9 \pm 0.8$ & \cellcolor{red!12}$29.9 \pm 0.9$ & \cellcolor{red!15}\xmark & \cellcolor{red!12}$40.0 \pm 8.7$ & $40.7 \pm 2.9$ & \cellcolor{red!12}$28.2 \pm 0.7$ & \cellcolor{red!12}$31.4 \pm 0.5$ & \cellcolor{red!15}\xmark \\
			\micon{intervl.png}~\mname{InternVL3.5-4B} & \cellcolor{red!12}$33.3 \pm 5.8$ & $48.0 \pm 2.0$ & $40.8 \pm 2.5$ & \cellcolor{red!12}$41.8 \pm 1.1$ & \cellcolor{red!15}\xmark & $43.3 \pm 2.9$ & $43.5 \pm 2.6$ & $41.2 \pm 1.9$ & \cellcolor{red!12}$41.8 \pm 1.6$ & \cellcolor{red!15}\xmark \\
			\micon{intervl.png}~\mname{InternVL3.5-8B} & \cellcolor{red!12}$30.0 \pm 0.0$ & $53.1 \pm 7.1$ & $48.2 \pm 2.0$ & \cellcolor{red!12}$47.9 \pm 2.9$ & \cellcolor{red!15}\xmark & $41.7 \pm 2.9$ & $49.2 \pm 6.8$ & $51.6 \pm 4.3$ & \cellcolor{red!12}$50.5 \pm 4.0$ & \cellcolor{red!15}\xmark \\
			\micon{intervl.png}~\mname{InternVL3.5-14B} & \cellcolor{red!12}$36.7 \pm 2.9$ & $47.5 \pm 1.7$ & $55.9 \pm 0.9$ & \cellcolor{red!12}$52.8 \pm 0.2$ & \cellcolor{red!15}\xmark & $50.0 \pm 8.7$ & $58.2 \pm 4.3$ & $58.8 \pm 1.3$ & \cellcolor{red!12}$58.1 \pm 1.0$ & \cellcolor{red!15}\xmark \\
			\micon{intervl.png}~\mname{InternVL3.5-30B-A3B} & \cellcolor{red!12}$33.3 \pm 2.9$ & $55.9 \pm 1.7$ & $65.3 \pm 1.3$ & $61.1 \pm 0.7$ & \cellcolor{red!15}\xmark & $48.3 \pm 5.8$ & $64.4 \pm 3.4$ & $68.9 \pm 1.2$ & $66.7 \pm 1.8$ & \cellcolor{green!15}\cmark \\
			\micon{intervl.png}~\mname{InternVL3.5-38B} & \cellcolor{red!12}$36.7 \pm 5.8$ & $53.7 \pm 2.6$ & $67.7 \pm 0.8$ & $62.5 \pm 0.7$ & \cellcolor{red!15}\xmark & $55.0 \pm 5.0$ & $66.1 \pm 2.9$ & $68.7 \pm 2.3$ & $67.3 \pm 2.3$ & \cellcolor{green!15}\cmark \\
			\micon{qwen.pdf}~\mname{Qwen2.5-VL-7B-Instruct} & $41.7 \pm 2.9$ & $53.1 \pm 1.0$ & $41.3 \pm 0.8$ & \cellcolor{red!12}$43.8 \pm 0.7$ & \cellcolor{red!15}\xmark & $45.0 \pm 0.0$ & $59.9 \pm 2.0$ & $46.9 \pm 0.4$ & \cellcolor{red!12}$49.3 \pm 0.2$ & \cellcolor{red!15}\xmark \\
			\micon{qwen.pdf}~\mname{Qwen2.5-VL-32B-Instruct} & $45.0 \pm 0.0$ & $55.9 \pm 0.0$ & $59.7 \pm 0.3$ & \cellcolor{red!12}$57.9 \pm 0.2$ & \cellcolor{red!15}\xmark & $51.7 \pm 2.9$ & $67.8 \pm 0.0$ & $67.1 \pm 0.0$ & $66.2 \pm 0.2$ & \cellcolor{green!15}\cmark \\
			\micon{qwen.pdf}~\mname{Qwen3-VL-2B-Instruct} & \cellcolor{red!12}$36.7 \pm 2.9$ & \cellcolor{red!12}$32.8 \pm 1.0$ & \cellcolor{red!12}$32.8 \pm 0.8$ & \cellcolor{red!12}$33.1 \pm 0.6$ & \cellcolor{red!15}\xmark & $46.7 \pm 2.9$ & \cellcolor{red!12}$35.0 \pm 1.0$ & \cellcolor{red!12}$28.8 \pm 0.9$ & \cellcolor{red!12}$31.2 \pm 0.7$ & \cellcolor{red!15}\xmark \\
			\micon{qwen.pdf}~\mname{Qwen3-VL-2B-Thinking} & $50.0 \pm 5.0$ & $46.9 \pm 2.6$ & $42.9 \pm 0.8$ & \cellcolor{red!12}$44.2 \pm 0.9$ & \cellcolor{red!15}\xmark & \cellcolor{red!12}$35.0 \pm 10.0$ & $50.8 \pm 2.9$ & $42.5 \pm 0.9$ & \cellcolor{red!12}$43.6 \pm 0.6$ & \cellcolor{red!15}\xmark \\
			\micon{qwen.pdf}~\mname{Qwen3-VL-4B-Instruct} & \cellcolor{red!12}$36.7 \pm 5.8$ & $42.9 \pm 2.6$ & $44.7 \pm 0.6$ & \cellcolor{red!12}$43.8 \pm 0.6$ & \cellcolor{red!15}\xmark & \cellcolor{red!12}$33.3 \pm 2.9$ & $58.2 \pm 2.0$ & $47.7 \pm 1.3$ & \cellcolor{red!12}$48.8 \pm 1.3$ & \cellcolor{red!15}\xmark \\
			\micon{qwen.pdf}~\mname{Qwen3-VL-4B-Thinking} & $53.3 \pm 7.6$ & $66.7 \pm 2.6$ & $68.6 \pm 3.6$ & $67.1 \pm 2.6$ & \cellcolor{green!15}\cmark & \cellcolor{red!12}$38.3 \pm 2.9$ & $71.2 \pm 2.9$ & $71.1 \pm 2.3$ & $68.9 \pm 1.7$ & \cellcolor{red!15}\xmark \\
			\micon{qwen.pdf}~\mname{Qwen3-VL-8B-Instruct} & \cellcolor{red!12}$36.7 \pm 5.8$ & $54.8 \pm 1.0$ & $56.9 \pm 0.9$ & \cellcolor{red!12}$55.0 \pm 0.9$ & \cellcolor{red!15}\xmark & \cellcolor{red!12}$36.7 \pm 2.9$ & $58.8 \pm 2.0$ & $53.7 \pm 0.3$ & \cellcolor{red!12}$53.5 \pm 0.5$ & \cellcolor{red!15}\xmark \\
			\micon{qwen.pdf}~\mname{Qwen3-VL-8B-Thinking} & $50.0 \pm 5.0$ & $74.0 \pm 1.0$ & $77.9 \pm 2.0$ & $75.1 \pm 2.0$ & \cellcolor{green!15}\cmark & $45.0 \pm 8.7$ & $83.1 \pm 1.7$ & $80.6 \pm 1.3$ & $78.7 \pm 1.3$ & \cellcolor{green!15}\cmark \\
			\micon{qwen.pdf}~\mname{Qwen3-VL-30B-A3B-Instruct} & \cellcolor{red!12}$38.3 \pm 5.8$ & $65.5 \pm 2.6$ & $72.3 \pm 1.3$ & $68.4 \pm 0.9$ & \cellcolor{red!15}\xmark & $46.7 \pm 2.9$ & $69.5 \pm 0.0$ & $74.3 \pm 1.4$ & $71.6 \pm 1.0$ & \cellcolor{green!15}\cmark \\
			\micon{qwen.pdf}~\mname{Qwen3-VL-30B-A3B-Thinking} & $45.0 \pm 5.0$ & $76.3 \pm 1.7$ & $84.7 \pm 1.8$ & $80.1 \pm 1.8$ & \cellcolor{green!15}\cmark & $48.3 \pm 5.8$ & $82.5 \pm 1.0$ & $86.0 \pm 0.4$ & $82.8 \pm 0.2$ & \cellcolor{green!15}\cmark \\
			\micon{qwen.pdf}~\mname{Qwen3-VL-32B-Instruct} & $45.0 \pm 5.0$ & $71.8 \pm 2.0$ & $74.6 \pm 1.0$ & $71.9 \pm 1.3$ & \cellcolor{green!15}\cmark & $48.3 \pm 2.9$ & $76.3 \pm 0.0$ & $83.8 \pm 0.4$ & $80.0 \pm 0.2$ & \cellcolor{green!15}\cmark \\
			\micon{qwen.pdf}~\mname{Qwen3-VL-32B-Thinking} & $48.3 \pm 12.6$ & $87.6 \pm 1.0$ & $83.5 \pm 1.2$ & $81.9 \pm 1.6$ & \cellcolor{green!15}\cmark & $58.3 \pm 2.9$ & $86.4 \pm 4.5$ & $88.9 \pm 1.3$ & $86.4 \pm 1.0$ & \cellcolor{green!15}\cmark \\
			\micon{gemini.png}~\mname{Gemma-3-4B-IT} & $40.0 \pm 0.0$ & \cellcolor{red!12}$35.6 \pm 0.0$ & \cellcolor{red!12}$33.0 \pm 0.3$ & \cellcolor{red!12}$34.0 \pm 0.2$ & \cellcolor{red!15}\xmark & \cellcolor{red!12}$35.0 \pm 0.0$ & \cellcolor{red!12}$27.1 \pm 0.0$ & \cellcolor{red!12}$32.7 \pm 0.3$ & \cellcolor{red!12}$31.8 \pm 0.2$ & \cellcolor{red!15}\xmark \\
			\micon{gemini.png}~\mname{Gemma-3-12B-IT} & $45.0 \pm 0.0$ & $61.0 \pm 0.0$ & $60.2 \pm 0.3$ & \cellcolor{red!12}$59.3 \pm 0.2$ & \cellcolor{red!15}\xmark & $50.0 \pm 0.0$ & $57.6 \pm 0.0$ & $61.0 \pm 0.0$ & \cellcolor{red!12}$59.6 \pm 0.0$ & \cellcolor{red!15}\xmark \\
			\micon{gemini.png}~\mname{Gemma-3-27B-IT} & $70.0 \pm 0.0$ & $59.3 \pm 0.0$ & $58.4 \pm 0.0$ & \cellcolor{red!12}$59.4 \pm 0.0$ & \cellcolor{red!15}\xmark & $45.0 \pm 0.0$ & $61.0 \pm 0.0$ & $57.7 \pm 0.3$ & \cellcolor{red!12}$57.5 \pm 0.2$ & \cellcolor{red!15}\xmark \\
			\micon{moonshot.pdf}~\mname{Kimi-VL-A3B-Instruct} & \cellcolor{red!12}$31.7 \pm 5.8$ & \cellcolor{red!12}$32.8 \pm 2.6$ & \cellcolor{red!12}$28.7 \pm 2.6$ & \cellcolor{red!12}$29.8 \pm 1.8$ & \cellcolor{red!15}\xmark & \cellcolor{red!12}$28.3 \pm 5.8$ & \cellcolor{red!12}$29.9 \pm 2.0$ & \cellcolor{red!12}$27.3 \pm 1.8$ & \cellcolor{red!12}$27.9 \pm 2.1$ & \cellcolor{red!15}\xmark \\
			\micon{moonshot.pdf}~\mname{Kimi-VL-A3B-Thinking} & $57.5 \pm 3.5$ & $40.7 \pm 7.2$ & $40.1 \pm 0.7$ & \cellcolor{red!12}$41.5 \pm 1.3$ & \cellcolor{red!15}\xmark & \cellcolor{red!12}$27.5 \pm 10.6$ & $42.4 \pm 2.4$ & \cellcolor{red!12}$36.2 \pm 0.9$ & \cellcolor{red!12}$36.8 \pm 0.5$ & \cellcolor{red!15}\xmark \\
			\micon{zai.pdf}~\mname{GLM-4.6V} & $55.0 \pm 10.0$ & $77.4 \pm 3.5$ & $82.2 \pm 1.0$ & $79.2 \pm 1.2$ & \cellcolor{green!15}\cmark & $50.0 \pm 0.0$ & $86.4 \pm 2.9$ & $87.7 \pm 0.9$ & $85.0 \pm 1.2$ & \cellcolor{green!15}\cmark \\
			\micon{zai.pdf}~\mname{GLM-4.6V-Flash} & $46.7 \pm 12.6$ & $70.6 \pm 2.0$ & $67.8 \pm 0.5$ & $66.9 \pm 0.7$ & \cellcolor{green!15}\cmark & \cellcolor{red!12}$38.3 \pm 2.9$ & $70.6 \pm 8.4$ & $67.3 \pm 0.5$ & $66.0 \pm 1.6$ & \cellcolor{red!15}\xmark \\
			\micon{mistral.png}~\mname{Ministral-3-3B-Instruct-2512} & \cellcolor{red!12}$38.3 \pm 2.9$ & $51.4 \pm 1.0$ & $41.6 \pm 3.0$ & \cellcolor{red!12}$43.4 \pm 2.5$ & \cellcolor{red!15}\xmark & $53.3 \pm 2.9$ & $49.2 \pm 3.4$ & $41.5 \pm 0.7$ & \cellcolor{red!12}$43.8 \pm 1.0$ & \cellcolor{red!15}\xmark \\
			\micon{mistral.png}~\mname{Ministral-3-3B-Reasoning-2512} & $50.0 \pm 10.0$ & $58.2 \pm 6.4$ & $49.8 \pm 1.6$ & \cellcolor{red!12}$51.6 \pm 1.6$ & \cellcolor{red!15}\xmark & $41.7 \pm 7.6$ & $58.8 \pm 1.0$ & $48.2 \pm 0.4$ & \cellcolor{red!12}$49.8 \pm 0.9$ & \cellcolor{red!15}\xmark \\
			\micon{mistral.png}~\mname{Ministral-3-8B-Instruct-2512} & $43.3 \pm 5.8$ & $58.2 \pm 1.0$ & $52.5 \pm 1.3$ & \cellcolor{red!12}$53.0 \pm 0.6$ & \cellcolor{red!15}\xmark & \cellcolor{red!12}$31.7 \pm 2.9$ & $58.8 \pm 2.6$ & $59.5 \pm 2.0$ & \cellcolor{red!12}$57.5 \pm 1.7$ & \cellcolor{red!15}\xmark \\
			\micon{mistral.png}~\mname{Ministral-3-8B-Reasoning-2512} & $51.7 \pm 2.9$ & $68.9 \pm 3.9$ & $66.3 \pm 3.0$ & $65.8 \pm 1.6$ & \cellcolor{green!15}\cmark & $41.7 \pm 5.8$ & $73.4 \pm 1.0$ & $69.3 \pm 1.8$ & $68.3 \pm 1.8$ & \cellcolor{green!15}\cmark \\
			\micon{mistral.png}~\mname{Ministral-3-14B-Instruct-2512} & $46.7 \pm 7.6$ & $69.5 \pm 2.9$ & $55.3 \pm 1.6$ & \cellcolor{red!12}$57.7 \pm 1.2$ & \cellcolor{red!15}\xmark & $58.3 \pm 7.6$ & $66.1 \pm 2.9$ & $62.1 \pm 2.2$ & $62.6 \pm 1.5$ & \cellcolor{green!15}\cmark \\
			\micon{mistral.png}~\mname{Ministral-3-14B-Reasoning-2512} & $45.0 \pm 5.0$ & $74.6 \pm 4.5$ & $78.2 \pm 1.8$ & $75.1 \pm 1.9$ & \cellcolor{green!15}\cmark & $43.3 \pm 10.4$ & $67.2 \pm 5.2$ & $79.4 \pm 2.4$ & $74.7 \pm 2.7$ & \cellcolor{green!15}\cmark \\
			\midrule
			\multicolumn{11}{l}{\emph{Medical-specific VLMs (n=9; pass: 1/9 in 2022, 3/9 in 2023)}} \\
			\midrule
			\micon{gemini.png}~\mname{MedGemma-1.5-4B-IT} & $51.7 \pm 2.9$ & $44.6 \pm 1.0$ & $50.0 \pm 1.3$ & \cellcolor{red!12}$49.0 \pm 0.8$ & \cellcolor{red!15}\xmark & $43.3 \pm 5.8$ & $42.9 \pm 5.2$ & $49.3 \pm 1.3$ & \cellcolor{red!12}$47.7 \pm 1.0$ & \cellcolor{red!15}\xmark \\
			\micon{gemini.png}~\mname{MedGemma-4B-IT} & $40.0 \pm 0.0$ & $53.7 \pm 1.0$ & \cellcolor{red!12}$36.1 \pm 0.0$ & \cellcolor{red!12}$40.1 \pm 0.2$ & \cellcolor{red!15}\xmark & \cellcolor{red!12}$35.0 \pm 0.0$ & \cellcolor{red!12}$35.6 \pm 0.0$ & \cellcolor{red!12}$38.2 \pm 0.0$ & \cellcolor{red!12}$37.5 \pm 0.0$ & \cellcolor{red!15}\xmark \\
			\micon{gemini.png}~\mname{MedGemma-27B-IT} & $50.0 \pm 0.0$ & $59.3 \pm 0.0$ & $58.6 \pm 0.3$ & \cellcolor{red!12}$58.1 \pm 0.2$ & \cellcolor{red!15}\xmark & $45.0 \pm 0.0$ & $59.3 \pm 0.0$ & $60.8 \pm 0.3$ & \cellcolor{red!12}$59.5 \pm 0.2$ & \cellcolor{red!15}\xmark \\
			\micon{lingshu_big.png}~\mname{Lingshu-7B} & $40.0 \pm 0.0$ & $52.5 \pm 0.0$ & $43.6 \pm 0.5$ & \cellcolor{red!12}$45.2 \pm 0.4$ & \cellcolor{red!15}\xmark & $61.7 \pm 2.9$ & $54.8 \pm 1.0$ & $49.7 \pm 0.3$ & \cellcolor{red!12}$51.5 \pm 0.3$ & \cellcolor{red!15}\xmark \\
			\micon{lingshu_big.png}~\mname{Lingshu-32B} & $40.0 \pm 0.0$ & $60.5 \pm 2.0$ & $65.7 \pm 0.3$ & $62.8 \pm 0.2$ & \cellcolor{green!15}\cmark & $45.0 \pm 0.0$ & $70.6 \pm 1.0$ & $69.3 \pm 0.0$ & $68.0 \pm 0.2$ & \cellcolor{green!15}\cmark \\
			\micon{hulu_med_big.png}~\mname{Hulu-Med-4B} & $40.0 \pm 0.0$ & $57.6 \pm 5.1$ & $48.2 \pm 1.5$ & \cellcolor{red!12}$49.6 \pm 2.1$ & \cellcolor{red!15}\xmark & $41.7 \pm 5.8$ & $47.5 \pm 5.1$ & $48.5 \pm 0.7$ & \cellcolor{red!12}$47.9 \pm 1.5$ & \cellcolor{red!15}\xmark \\
			\micon{hulu_med_big.png}~\mname{Hulu-Med-7B} & $48.3 \pm 7.6$ & $52.0 \pm 2.6$ & $47.2 \pm 1.0$ & \cellcolor{red!12}$48.3 \pm 0.8$ & \cellcolor{red!15}\xmark & \cellcolor{red!12}$36.7 \pm 2.9$ & $52.0 \pm 1.0$ & $51.9 \pm 1.8$ & \cellcolor{red!12}$50.9 \pm 1.5$ & \cellcolor{red!15}\xmark \\
			\micon{hulu_med_big.png}~\mname{Hulu-Med-14B} & $41.7 \pm 2.9$ & $53.7 \pm 7.1$ & $62.0 \pm 2.5$ & \cellcolor{red!12}$58.8 \pm 2.4$ & \cellcolor{red!15}\xmark & $43.3 \pm 5.8$ & $60.5 \pm 4.3$ & $65.4 \pm 0.4$ & $63.0 \pm 1.5$ & \cellcolor{green!15}\cmark \\
			\micon{hulu_med_big.png}~\mname{Hulu-Med-32B} & \cellcolor{red!12}$38.3 \pm 2.9$ & $68.9 \pm 2.0$ & $63.4 \pm 1.3$ & $62.8 \pm 1.1$ & \cellcolor{red!15}\xmark & $45.0 \pm 5.0$ & $66.7 \pm 2.0$ & $73.7 \pm 2.7$ & $70.5 \pm 1.8$ & \cellcolor{green!15}\cmark \\
			\midrule
			\multicolumn{11}{l}{\emph{Korean-specific VLMs (n=5; pass: 0/5 in 2022, 0/5 in 2023)}} \\
			\midrule
			\micon{nc.png}~\mname{VARCO-VISION-2.0-1.7B} & $45.0 \pm 0.0$ & \cellcolor{red!12}$37.3 \pm 0.0$ & \cellcolor{red!12}$29.2 \pm 0.0$ & \cellcolor{red!12}$32.0 \pm 0.0$ & \cellcolor{red!15}\xmark & $45.0 \pm 0.0$ & \cellcolor{red!12}$33.9 \pm 0.0$ & \cellcolor{red!12}$28.5 \pm 0.0$ & \cellcolor{red!12}$30.6 \pm 0.0$ & \cellcolor{red!15}\xmark \\
			\micon{nc.png}~\mname{VARCO-VISION-2.0-14B} & \cellcolor{red!12}$30.0 \pm 0.0$ & $55.9 \pm 0.0$ & $54.0 \pm 0.0$ & \cellcolor{red!12}$52.7 \pm 0.0$ & \cellcolor{red!15}\xmark & $40.0 \pm 0.0$ & $54.2 \pm 0.0$ & $51.8 \pm 0.0$ & \cellcolor{red!12}$51.5 \pm 0.0$ & \cellcolor{red!15}\xmark \\
			\micon{skt.png}~\mname{A.X-4.0-VL-Light} & $40.0 \pm 5.0$ & $58.2 \pm 3.5$ & $46.7 \pm 2.1$ & \cellcolor{red!12}$48.6 \pm 1.6$ & \cellcolor{red!15}\xmark & $58.3 \pm 2.9$ & $50.3 \pm 2.6$ & $47.5 \pm 1.8$ & \cellcolor{red!12}$48.8 \pm 1.1$ & \cellcolor{red!15}\xmark \\
			\micon{naver.png}~\mname{HyperCLOVAX-SEED-Vision-Instruct-3B} & \cellcolor{red!12}$35.0 \pm 0.0$ & \cellcolor{red!12}$32.2 \pm 0.0$ & \cellcolor{red!12}$30.7 \pm 0.5$ & \cellcolor{red!12}$31.3 \pm 0.4$ & \cellcolor{red!15}\xmark & \cellcolor{red!12}$35.0 \pm 0.0$ & \cellcolor{red!12}$31.6 \pm 1.0$ & \cellcolor{red!12}$28.2 \pm 1.1$ & \cellcolor{red!12}$29.3 \pm 0.7$ & \cellcolor{red!15}\xmark \\
			\micon{kakao.png}~\mname{kanana-1.5-v-3b-instruct} & $40.0 \pm 0.0$ & $40.7 \pm 0.0$ & \cellcolor{red!12}$31.2 \pm 0.0$ & \cellcolor{red!12}$33.8 \pm 0.0$ & \cellcolor{red!15}\xmark & $50.0 \pm 0.0$ & \cellcolor{red!12}$35.6 \pm 0.0$ & \cellcolor{red!12}$37.3 \pm 0.0$ & \cellcolor{red!12}$37.8 \pm 0.0$ & \cellcolor{red!15}\xmark \\
			\midrule
			\multicolumn{11}{l}{\textit{Summary: 15/51 models pass in 2022, 20/51 in 2023}} \\
			\bottomrule
		\end{tabular}%
	}
\end{table}

\end{document}